\PassOptionsToPackage{table}{xcolor}
\documentclass{article}
\usepackage[final]{colm2025_conference}

\usepackage{microtype}
\usepackage{url}
\usepackage{booktabs}
\usepackage{lineno}

\usepackage{import}

\usepackage[T1]{fontenc}
\usepackage[utf8]{inputenc}
\usepackage{graphicx}
\usepackage{comment}
\usepackage{enumitem}
\usepackage{caption}
\usepackage{subcaption}
\usepackage{float}
\usepackage{tikz}
\usepackage{pifont}
\usepackage{xspace}

\usepackage{hyperref}
\definecolor{darkblue}{rgb}{0, 0, 0.5}
\hypersetup{colorlinks=true, citecolor=darkblue, linkcolor=darkblue, urlcolor=darkblue}

\newcommand{\citeTODO}{\textcolor{red}{[CITE?]}%
\gappto{\citetodolist}{\protect\noindent Line \the\inputlineno: TODO Citation Needed \thesection.\\}}
\newcommand{\citetodolist}{}

\newcommand\refsec[1]{\hyperlink{#1}{§\ref{sec:#1}:~\textsc{#1}}}

\newcommand\reffig[1]{Figure~\ref{fig:#1}}

\newcommand\reffigss[3]{Figures~\ref{fig:#1},~\ref{fig:#2}, and~\ref{fig:#3}}

\newcommand\reftab[1]{Table~\ref{tab:#1}}
\newcommand\refapp[1]{Appendix~\ref{app:#1}}

\usepackage{amsmath,amsfonts,bm}

\def\eqref#1{equation~\ref{#1}}

\def\1{\bm{1}}

\DeclareMathAlphabet{\mathsfit}{\encodingdefault}{\sfdefault}{m}{sl}
\SetMathAlphabet{\mathsfit}{bold}{\encodingdefault}{\sfdefault}{bx}{n}

\title{Do Language Models Agree with Humans\\Perceptions of Suspense in Stories?}
\author{Glenn Matlin\thanks{:Corresponding author, \textsuperscript{$\diamond$}: Equal contribution, \texttt{https://github.com/eilab-gt/SuspensePerception}}, Devin Zhang\textsuperscript{$\diamond$}, Rodrigo Loza\textsuperscript{$\diamond$}, Diana M. Popescu\textsuperscript{$\diamond$},~\\Jacqueline Isbell, Chandreyi Chakraborty, Mark Riedl
\\
School of Interactive Computing, College of Computing\\
Georgia Institute of Technology, USA \\
\\
\small{
    \textbf{Correspondence:} \href{mailto:glenn@gatech.edu}{glenn@gatech.edu}}
}
\begin{document}

\ifcolmsubmission
\linenumbers
\fi

\maketitle
\begin{abstract}
Suspense is an affective response to narrative text that is believed to involve complex cognitive processes in humans.
Several psychological models have been developed to describe this phenomenon and the circumstances under which text might trigger it.
We replicate four seminal psychological studies of human perceptions of suspense, substituting human responses with those of different open-weight and closed-source LMs. 
We conclude that while LMs can distinguish whether a text is intended to induce suspense in people,
LMs cannot accurately estimate the relative amount of suspense within a text sequence as compared to human judgments,
nor can LMs properly capture the human perception for the rise and fall of suspense across multiple text segments.
We examine the functional limits of LMs suspense understanding capabilities by adversarially permuting the story text to probe what causes the perception of suspense to diverge for both human and LMs. 
We conclude that, while LMs can superficially identify and track certain facets of suspense, they do not process suspense in the same way as human readers.
\end{abstract}
\section{Introduction}\label{sec:introduction}
Suspense, a multifaceted emotional phenomenon of the human mind, has been extensively studied across psychology~\citep{Gerrig1994-et,Madrigal2011-dx}, cognitive science~\citep{Colby1989-ex,Hoeken2000-jp} literary analysis~\citep{Brewer1982-ml,Brewer1996-rb}, and film studies~\citep{branigan92, Carroll1984-zw, Vorderer1996-xj}. In storytelling---a fundamental medium for communication, entertainment, education, conveying intricate ideas, building rapport, and social commentary---suspense plays a pivotal role by engaging audiences in complex emotional states \citep{Gerrig1994-et}.
Stories capable of generating suspense can heighten a reader's emotional investment, making them more likely to connect deeply with the content \citep{Doust2017-ut, O-Neill2014-ta}.
Whether in games \citep{Li2021-di}, news \citep{Kaspar2016-bl}, sports~\citep{Peterson2008-tp}, or business \citep{Hsieh2021-ef, Guidry2005-wq, Moulard2012-pi, Palich1995-ao}, the strategic evocation of suspense shapes individuals' emotional experiences and decision-making processes. By amplifying tension and creating unresolved questions, suspense compels readers or viewers to continuously seek resolution.
Suspense is believed to be an affective response to narrative content, written, visual, or otherwise, that involves complex cognitive phenomena such as empathy, perceptions of uncertainty, and anticipation~\citep{Gerrig1994-et}.

Suspense has also been a topic of study in Artificial Intelligence (AI), including the detection of suspense in narrative texts~\citep{Fendt2017LeveragingIR,Delatorre2020PredictingTE,Delatorre2021ImprovingTF,oneill:acii2011,O-Neill2014-ta,Doust2017-ut} and the generation of texts that evoke suspense in human readers~\citep{Fendt2017LeveragingIR,Delatorre2020PredictingTE,Delatorre2021ImprovingTF,Xie2024-qv}
Furthermore, AI models that can detect sophisticated narrative devices such as suspense may be used to develop intelligent human creativity support tools. While writing support is often approached from a generative standpoint, constructive critique can also support human creators~\citep{Lin2023Beyond}. Creative writing therefore serves as a useful benchmark for testing the capabilities of LMs~\citep{Paech2023-bt} and AI agents utilizing LMs~\citep{Gooding2025-tp}
In this paper we concern ourselves with the {\bf detection} and {\bf rating} of suspense in narrative text.
Early attempts at computationally modeling suspense relied on symbolic representations~\citep{Cheong2015SuspenserAS,oneill:acii2011, O-Neill2014-ta, Doust2017-ut}.
However, these techniques required manual knowledge engineering to convert stories into symbolic representations for analysis.
Language Models (LMs) based on the Transformer architecture \citep{Vaswani2017-qo} have shown remarkable capabilities in tasks such as sentiment analysis, summarization, and question answering. 
The question of whether LMs can make judgments of story suspense that correspond to human perceptions is not necessarily straightforward.
Suspense is a \emph{complex affective phenomena} potentially involving empathy, perceptions of uncertainty, and prediction of outcomes.
This complexity has meant that LMs are not particularly adept at generating suspenseful narratives~\citep{Xie2024-qv}. 
Generation is not the same as detection, which leads to our overall research question: how does the LM rating of suspense in written narrative texts relate to human perceptions of suspense in the same stories?

Our study investigates the ability of current LMs to detect and measure suspense in textual passages. Specifically, we pose four research questions aimed at probing how closely LMs can mirror human suspense perception:
\textbf{(1)}~Can LMs distinguish whether a text sequence is intended to generate suspense?
\textbf{(2)}~Do LMs accurately estimate the \emph{relative} amount of suspense within a text sequence compared to human judgments?
\textbf{(3)}~Can LMs identify the key moments when story suspense \emph{rises and falls} across multiple text segments, in alignment with human ratings?
\textbf{(4)}~Does adversarially permuting story text cause a divergence between human and LM perceptions of the text?
We conduct a \textit{machine psychology} study by replicating four notable psychology studies of human perceptions of suspense by swapping LMs for human participants. We design our experiment to follow the best practices in machine psychology \citep{Lohn2024-jg}. We investigate multiple families of LMs at a variety of parameter sizes, both open and closed source. We compare LM results to the data from the human participant studies.

Our findings lead us to conclude that while LMs can superficially identify and even track certain facets of suspense, they do not process or represent suspense in the same way human readers do. LMs struggle to quantify suspense levels on a continuous scale, and the machine's response often fails to accurately align with human judgments. LMs stumble even when identifying the most critical moments of rising and falling action across a narrative. Our experiments with adversarial or context-disrupting permutations show that LMs do not significantly alter their generated answers, even when the attacked text is significantly altered to increase/decrease suspense. Overall, we conclude that the surveyed LMs do ``mimic'' suspense judgments to a certain extent, but the manner in which they do so is sufficiently brittle---and often context-insensitive---that the measures made by these LMs diverge from the dynamic, human experience of unfolding tension.
\section{Related Work}\label{sec:related_work}
Understanding whether LMs can detect suspense falls within the domain of \textit{machine psychology}~\citep{Hagendorff2023MachinePI}, which applies psychological experiments, originally designed to study human cognition and behavior, to investigate computational models of intelligence and behavior. 
Several notable studies have directly examined and quantified human perceptions of suspense in stories, yielding alternative explanations for how the feeling of suspense arises from reading. 
\citet{Gerrig1994-zo} models suspense as a function of an individual's assessment of the possible solutions within a given problem and the complexity of transitioning from an initial state to a desired outcome. 
\citet{Wilmot2020-qa} explores these transitional states, proposing that a hierarchical language model leveraging uncertainty reduction can closely match human annotations of suspense in short stories.
\citet{Lehne2015-jw} proposes a cognitive model of suspense that conceptualizes it as an affective state driven primarily by uncertainty and anticipation about future narrative outcomes.
\citet{Brewer1988-va} developed Structural-Affect Theory, positing that suspense arises specifically when an initiating narrative event generates emotional uncertainty by delaying the resolution or outcome.
\citet{Delatorre2018-fl} measures the effects of uncertainty on suspense perception by providing one group of participants the ending in advance. \citet{Brewer1988-va} tests the Structural-Affect Theory, hypothesizing that the structural feature of an initiating event and the inclusion of its outcome produce suspense. 
\citet{Haider2020-tr} tests language model BERT on emotion classification, including suspense, using their annotated dataset on poetry. They found suspense was among the emotions the model had the most difficulty predicting.
\citet{Bentz2024-qu} produces a novel experiment to rate the suspense arc of a narrative in real time. As participants read a story, they drew a line that correlated to their perceived suspense levels.

From these and other papers, we drew our datasets based on the requirement that the papers could be replicated under the machine psychology experimental conditions suggested by~\citep{Lohn2024-jg} and provided quantifiable data in the form of human-annotated ratings of textual suspense on a Likert scale. From this criteria we arrived at our specific experiment dataset using results from \citet{Gerrig1994-zo}, \citet{Brewer1988-va}, \citet{Lehne2015-jw}, and \citet{Delatorre2018-fl}.

Sentiment analysis \citep{Liu2012-sy, Wankhade2022-oi} and emotional analysis \citep{Plaza-del-Arco2024-kx} attempt to classify text in terms of whether the author is attempting to convey a sentiment or whether a character is in a particular emotional state \citep{Zhu2011-xq}.
While suspense is also an affective response, suspense differs from the above in that suspense is the emotion that the {\em reader} will experience upon consuming narrative content.

Early work aimed at computationally model and detect suspense used symbolic logical systems~\citep{Cheong2015SuspenserAS,oneill:acii2011,O-Neill2014-ta,Doust2017-ut}.
\citet{Steg2022-lx} trained machine learning models to classify suspense in texts using reader annotations. They collected annotations on suspense, curiosity, and surprise, as well as human ratings from \citet{Piper2021-ql}, who assessed situatedness, event sequencing, and world-making.
\citet{Wilmot2020-qa} analyzed suspense in short stories by computing vector embeddings and measuring uncertainty reduction through a forward-looking probability distribution of possible narrative paths. This method proved to be a strong predictor of human-perceived suspense, as it represented when resolution appeared imminent.
Our work directly leverages LLMs for suspense detection rather than traditional NLP models, benefiting from their ability to capture long-range dependencies in text and training on large corpora.
\section{Experimental Setup}
\label{sec:experimental-setup}

In this section, we detail our methodology for evaluating Large Language Models (LMs) on the task of suspense detection. 
By adapting classic psychology experiments to a modern AI context, we explore whether models can mirror the same rise, peak, and resolution of suspense that human readers experience.
To conduct a \textit{machine psychology} experiment, we follow the recommendations and requirements developed in \cite{Hagendorff2023MachinePI} for how to reduce bias in machine psychology studies and ensure a fair comparison across models and with humans.

\subsection{Dataset and Sources}
\label{subsec:data_sources}
We replicate four foundational psychology studies to capture the diverse ways in which suspense manifests. These works vary in text length, genre, and annotation style, allowing us to test whether LMs can generalize across narrative structures and psychological instruments.
Each employed rigorous methodologies to obtain human suspense ratings. 
Selection criteria prioritized datasets that offered demographic diversity and variation in text segment length, ensuring a broader representation of suspense perception. We specifically selected the studies that utilized a Likert scale. This choice enables a quantifiable comparison between human ratings and model predictions. 

\textbf{\citet{Gerrig1994-zo}} used excerpts from a James Bond novel, originally rated by participants at Yale University and the University of the Philippines on two dimensions: \emph{likelihood of escape} and \emph{suspense}. Each excerpt was assigned a numeric rating on a 1--7 scale, reflecting the intensity of participants’ reactions. We maintain the experimental conditions (e.g., presence or absence of a key item) and replicate these scenarios for the LM prompts.

\textbf{\citet{Brewer1988-va}} introduced a structural-affect theory of stories, using American and Hungarian short stories. Participants rated suspense at five key plot segments for each story. This segmentation is ideal for our multi-step model prompts, as it requires the LM to \emph{incrementally} track how suspense evolves as the narrative unfolds.

\textbf{\citet{Delatorre2018-fl}}  investigated how the valence of the ending (good/bad), the revelation of the ending (revealed/unrevealed), and writing style (journalistic/novelistic) affect suspense in an English-language short story. We use the original twelve segments as published and preserve the original rating scale (1--9) The authors intentionally designed the story narrative used in the original experiment to follow the classical pyramidal structure proposed by \citet{Freytag1894-zz}, explicitly incorporating stages of exposition, rising action, climax, falling action, and denouement (see Figure ~\ref{fig:freytags_pyramid}). 
The rising action deliberately introduces uncertainty while the falling action explicitly resolves this suspense. 

\textbf{\citet{Lehne2015-jw}}  measured segment-level suspense in a shortened German-language version of \emph{`The Sandman'} (1816; German: \textit{Der Sandmann}) by E. T. A. Hoffmann. We used the publisher's official English translation divided into 65 segments matching the original German experiment. The human participants rated each segment in real time. The question is whether LMs similarly track the continuous ``ebb and flow” of suspense across a longer narrative, aligning with human data.

\subsection{Replicating the Studies with LMs}
\label{subsec:incremental_reading}
We replicate each study’s instructions nearly verbatim, ensuring that the LM produces a numeric value within the specified range used in each original human study. 
If the original research collected multiple ratings along different dimensions (e.g., “likelihood of escape” and “suspense”), we prompt the model to provide both.
Brewer asks some additional questions along several dimensions beyond suspense (i.e., rating the segment's ironic connotation and its relation to the suspense); we did not replicate these questions.
Gerrig's participants were provided with entire passages of text before answering questions.
Brewer, Delatorre, and Lehne conducted their studies using a text sample that spanned multiple segments or the story. The resulting labels from human annotators from each study were provided at an aggregate level, limiting our ability to analyze inter-rater agreement or similar annotation quality metrics.
To replicate these studies, the machine receives each segment sequentially, which prevents LMs from attending to elements that will appear later in the story than each particular segment being asked about. The full experiment and story are conducted in a multi-message conversational format.  During this conversation, we provide the instructions (e.g., “Rate how suspenseful you find this passage on a 1--7 scale”) to every segment, staying in line with the requirements of a machine psychology experiment. We append the message and response from the first segment inquiry to the front of the second prompt and segment, continuing in this manner until the story is finished and the experiment is completed.

\subsection{LM Selection and Configurations}
\label{subsec:LM_selection}
We evaluate multiple state-of-the-art LMs that were available and widely used contemporaneously with our study. We selected these language models based on their high performance on numerous common language modeling leaderboards~\citep{Liang2022-ew, Wang2024-pk}. We select LMs to offer as much diversity as possible by selecting LMs that differ in parameter count, training corpus, and availability (closed vs.\ open source). To holistically evaluate suspense, we focus our evaluation on foundational instruction-tuned language models (LMs) and do not fine-tune models specifically for the suspense detection task. 
\textbf{open-weight models} include: Llama2 7B, Llama3 8B, Llama3 70B, Gemma2 9B, Gemma2 27B, DeepSeek-V3, Qwen2 72B, Mistral 7B, Mixtral 8x7B, and WizardLM2 8x22B. To minimize variability between samples, we set the \textit{temperature} to 0 for models where we can control the randomness of sampling. We repeat each experiment three times and average the outputs, while acknowledging the residual randomness of the model that we cannot directly seed.

\subsection{Evaluation Metrics}
\label{subsec:evaluation_metrics}
We aim to evaluate the extent to which LMs accurately capture the presence, relative magnitude, and trajectory of suspense.
All the experiment data from \citet{Gerrig1994-zo}, \cite{Brewer1988-va}, \cite{Delatorre2018-fl}, and \citet{Lehne2015-jw} used numerical ratings with a middle neutral point on a scale of either 1--7 or 1--9.
We measure the {\bf correlational error} using standard metrics between the model's generated rating value and the mean human rating value.
Mean Absolute Error (MAE) measures the distance between the LM predictions and the human average.

Recognizing that an LM may have a different understanding of numerical scales than humans and demonstrate specific biases favoring certain numbers or scale \citep{Shaki2023-eg, Shah2023-ej, Shaikh2024-bv, Malberg2024-bz}, we present results when we relax the requirement for the LM to directly predict the magnitude of the numeric value. We analyze performance when performing binarization the machine's generated values on the Likert scale into a binary label representing the level of ``high'' or ``low'' suspense. Using a binary classification, we provide analysis regarding whether the generated values by these LM would label the segments as having ``high'' or ``low'' suspense, mimicking how human participants suspense would be anchored to the midpoint separating the upper or lower halves of the Likert scale.

We also look at the {\bf directionality} of suspense ratings---whether suspense is increasing or decreasing from one segment of story to the next. We compute the above metrics on a per-segment level, we can also see where suspense perceptions are rising or falling and whether the LM's ratings are also rising or falling between the same pair of segments. 

Each model is run three times on each story segment, after which we take the mean rating per segment. 
We then calculate metrics using averaged outputs. This repetition partially offsets unknown sampling variance for closed-source models and ensures consistent reporting.

\section{Results}
\label{sec:results}

The results for rating agreement for Gerrig are in \reffig{gerrig-value}, which is provided as a heatmap.
Each column is a story treatment---Gerrig gives different stories with variations to increase or reduce suspense.
Each row is a different model, and the bottom row has the average human ratings.
Yellow indicates close match between model and human ratings, and blue indicates a large discrepancy (too high or too low).
The heatmap is designed to allow for visual inspection of the general trends of models and stories. 
\refapp{results} gives full-size and charts the details of the story variations.
Generally, the LMs vary significantly from human ratings, either reporting too much suspense or too little. 
\reftab{gerrig} shows agreement when we binarize the human data, with any rating above the middle rating indicating presence of suspense. 
The LMs are generally in agreement with humans when we binarize the data. 
We note that the story variations decrease suspense but never below the middle point, and the LMs report all stories suspenseful.
We also note that the Gerrig experiment uses James Bond stories, which are part of all LM pre-training data.

\reffigss{brewer-value}{delatorre-value}{lehne-value} show results for Brewer, Delatorre, and Lehne, respectively (Brewer had multiple stories and we only show the results for one story here; others can be found in \refapp{results}).
In these experiments, storys are broken into segments. 
Generally speaking, the results are a mix of close matches and misses. 
As before, full size heatmaps are in \refapp{results}.
Larger models tend to produce more accurate matches, but not uniformly so.
\reftab{brewer-delatorre-lehne} show matches on binarized data where we see agreement at 78\% or above.

Because Delatorre and Lehne have stories broken into segments and have sufficient length, we are able to conduct directionality analysis wherein we mark agreement if the segment-to-segment ratings are rising or falling in accordance with rising or falling human ratings. See \reffig{directionality} (top).
We see that model ratings trend in the same direction as human ratings, indicating that LMs can accurately assess the coarser notion of rising and falling suspense.
There are vertical bands of darker colors in the heatmaps that warrant further analysis. 
These bands correlate with a switch from rising suspense to falling suspense, or vice-versa, according to human ratings.
When we isolate and analyze these inflection points where rising changes to falling (\reffig{directionality} bottom), we see that LMs are substantially less reliable.
However, when we look at binarized scales (suspense present/not-present), agreement does not noticeably drop (see \refapp{results}).

Most significantly, LMs failed to detect suspense where humans succeeded for passages that increase suspense in the context of the story while being not inherently suspenseful on their own. Due to the fact that we fed the models text in chunks for Brewer, Delatorre, and Lehne with the repeated prompt, we considered that the LMs may be answering the prompt without considering how the text relates to the storyline as a whole and reran the LMs so the rating prompt ``Rate the following paragraph for its suspensefulness on a 9-point scale, where 1 is `not suspenseful' and 9 is `very suspenseful' '' included ``in relation to the storyline so far.'' at the end. However, we found that this did not significantly positively or negatively impact the results of the LMs.

\noindent
\begin{minipage}[htbp]{1.0\textwidth}%
  \centering
  \scriptsize
  \setlength{\tabcolsep}{3pt}
  \renewcommand{\arraystretch}{1}
  \begin{tabular}{l|cccc}
    \textbf{Model} & \textbf{Acc.} $\uparrow$ & \textbf{F1} $\uparrow$ & \textbf{MSE} $\downarrow$ \\
    \hline
    Qwen2 72B      & 1.00 $\pm$ 0.00 & 1.00 $\pm$ 0.00 & 13.38 $\pm$ 4.33 \\
    DeepSeek V3    & 1.00 $\pm$ 0.00 & 1.00 $\pm$ 0.00 & 13.38 $\pm$ 4.33 \\
    Gemma 2 27b    & 1.00 $\pm$ 0.00 & 1.00 $\pm$ 0.00 & 12.02 $\pm$ 4.82 \\
    Llama 2 7b     & 1.00 $\pm$ 0.00 & 1.00 $\pm$ 0.00 & 13.38 $\pm$ 4.33 \\
    Llama 3 70b    & 1.00 $\pm$ 0.00 & 1.00 $\pm$ 0.00 & 13.38 $\pm$ 4.33 \\
    Llama 3 8b     & 1.00 $\pm$ 0.00 & 1.00 $\pm$ 0.00 & 26.15 $\pm$ 7.15 \\
    WizardLM 2 8x22B & 1.00 $\pm$ 0.00 & 1.00 $\pm$ 0.00 & 13.38 $\pm$ 4.33\\
    Mistral 7B     & 1.00 $\pm$ 0.00 & 1.00 $\pm$ 0.00 & \cellcolor{yellow!25}7.84 $\pm$ 5.48 \\
    Mixtral 8x7B   & 1.00 $\pm$ 0.00 & 1.00 $\pm$ 0.00 & 8.87 $\pm$ 6.41 \\
    \hline
    Average        & 1.00 $\pm$ 0.00 & 1.00 $\pm$ 0.00 & 13.18 $\pm$ 5.21 \\
    \hline
  \end{tabular}
\captionof{table}{Gerrig results on binary yes/no scale.}
\label{tab:gerrig}
\end{minipage}
\setlength{\tabcolsep}{3pt}
\renewcommand{\arraystretch}{1}
\begin{table}[htbp]
\centering
\resizebox{\textwidth}{!}{\begin{tabular}{l|ccc|ccc|ccc}
&  \multicolumn{3}{c|}{\bf Brewer} & \multicolumn{3}{c|}{\bf Delatorre} & \multicolumn{3}{c}{\bf Lehne} \\
\hline
\textbf{Model} & \textbf{Acc.} $\uparrow$ & \textbf{F1} $\uparrow$ & \textbf{MSE} $\downarrow$  & \textbf{Acc.} $\uparrow$ & \textbf{F1} $\uparrow$ & \textbf{MSE} $\downarrow$  & \textbf{Acc.} $\uparrow$ & \textbf{F1} $\uparrow$ & \textbf{MSE} $\downarrow$ \\
\hline 
Qwen2 72B 
& \cellcolor{yellow!25}1.00 $\pm$ 0.00 & \cellcolor{yellow!25}1.00 $\pm$ 0.00 & 4.61 $\pm$ 2.09  
& 0.68 $\pm$ 0.10 & 0.77 $\pm$ 0.08 & 8.81 $\pm$ 1.68 
& 0.80 $\pm$ 0.00  & 0.88 $\pm$ 0.00  & 4.53 $\pm$ 0.00   
\\
DeepSeek V3 
& 0.93 $\pm$ 0.15 & 0.96 $\pm$ 0.09 & 8.08 $\pm$ 4.54 
& 0.69 $\pm$ 0.12 & 0.79 $\pm$ 0.09 & 6.54 $\pm$ 0.68 
& \cellcolor{yellow!25}0.86 $\pm$ 0.00 & 0.91 $\pm$ 0.00 & 4.17 $\pm$ 0.00  
\\
Gemma 2 27b 
& 0.80 $\pm$ 0.27 & 0.87 $\pm$ 0.18 & 5.31 $\pm$ 2.51 
& \cellcolor{yellow!25}0.78 $\pm$ 0.06 & \cellcolor{yellow!25}0.85 $\pm$ 0.04 &\cellcolor{yellow!25} 5.13 $\pm$ 0.79  
& \cellcolor{yellow!25}0.86 $\pm$ 0.00 & \cellcolor{yellow!25}0.92 $\pm$ 0.00 & 5.35 $\pm$ 0.00  
\\
Gemma 2 9b 
& 0.80 $\pm$ 0.33 & 0.85 $\pm$ 0.26 & 3.84 $\pm$ 2.77 
& 0.68 $\pm$ 0.08 & 0.80 $\pm$ 0.05  & 7.47 $\pm$ 2.11 
& 0.83 $\pm$ 0.00 & 0.90 $\pm$ 0.00 & 5.90 $\pm$ 0.00
\\
Llama 2 7b 
& 0.19 $\pm$ 0.38 & 0.21 $\pm$ 0.43 & 18.47 $\pm$ 14.97
& 0.68 $\pm$ 0.05 & 0.81 $\pm$ 0.04 & 12.19 $\pm$ 1.51 
& 0.83 $\pm$ 0.00 & 0.91 $\pm$ 0.00 & 12.95 $\pm$ 0.00 
\\
Llama 3 70b 
& 0.72 $\pm$ 0.31 & 0.80 $\pm$ 0.25 & 4.79 $\pm$ 3.74  
& 0.72 $\pm$ 0.11 & 0.79 $\pm$ 0.11 & 12.22 $\pm$ 1.91  
& 0.77 $\pm$ 0.00 & 0.84 $\pm$ 0.00 & 7.87 $\pm$ 0.00  
\\
Llama 3 8b 
& 0.92 $\pm$ 0.18 & 0.95 $\pm$ 0.11 & 1.95 $\pm$ 0.83  
& 0.65 $\pm$ 0.12 & 0.76 $\pm$ 0.11 & 9.92 $\pm$ 1.00  
& 0.72 $\pm$ 0.00 & 0.81 $\pm$ 0.00 & 8.83 $\pm$ 0.00  
\\
WizardLM 2 
& 0.80 $\pm$ 0.45 & 0.80 $\pm$ 0.45 & 6.88 $\pm$ 5.97  
& 0.70 $\pm$ 0.09 & 0.80 $\pm$ 0.07 & 7.85 $\pm$ 3.51 
& 0.83 $\pm$ 0.00 & 0.89 $\pm$ 0.00 & \cellcolor{yellow!25}4.09 $\pm$ 0.00  
\\
Mistral 7B 
& 0.76 $\pm$ 0.36 & 0.82 $\pm$ 0.29 & \cellcolor{yellow!25}1.52 $\pm$ 0.58 
& 0.68 $\pm$ 0.08 & 0.79 $\pm$ 0.07 & 8.27 $\pm$ 2.13  
& 0.85 $\pm$ 0.00 & 0.91 $\pm$ 0.00 & 10.58 $\pm$ 0.00  
\\
Mixtral 8x7B 
& 0.45 $\pm$ 0.37 & 0.55 $\pm$ 0.37 & 9.20 $\pm$ 10.61 
& 0.69 $\pm$ 0.07 & 0.80 $\pm$ 0.05 & 10.22 $\pm$ 2.01  
& 0.71 $\pm$ 0.00 & 0.79 $\pm$ 0.00 & 5.88 $\pm$ 0.00  
\\
\hline
Average 
& 0.74 $\pm$ 0.28 & 0.78 $\pm$ 0.24 & 6.46 $\pm$ 4.86  
& 0.69 $\pm$ 0.09 & 0.80 $\pm$ 0.07 & 8.86 $\pm$ 1.73  
& 0.81 $\pm$ 0.00 & 0.88 $\pm$ 0.00 & 7.02 $\pm$ 0.00  
\\
\hline
\end{tabular}
}
\caption{Brewer, Delatorre, and Lehne results on binary yes/no scale.}
\label{tab:brewer-delatorre-lehne}
\end{table}

\begin{figure}[t]
    \centering
    \begin{minipage}{1\textwidth}
        \centering
        \includegraphics[width=1\textwidth]{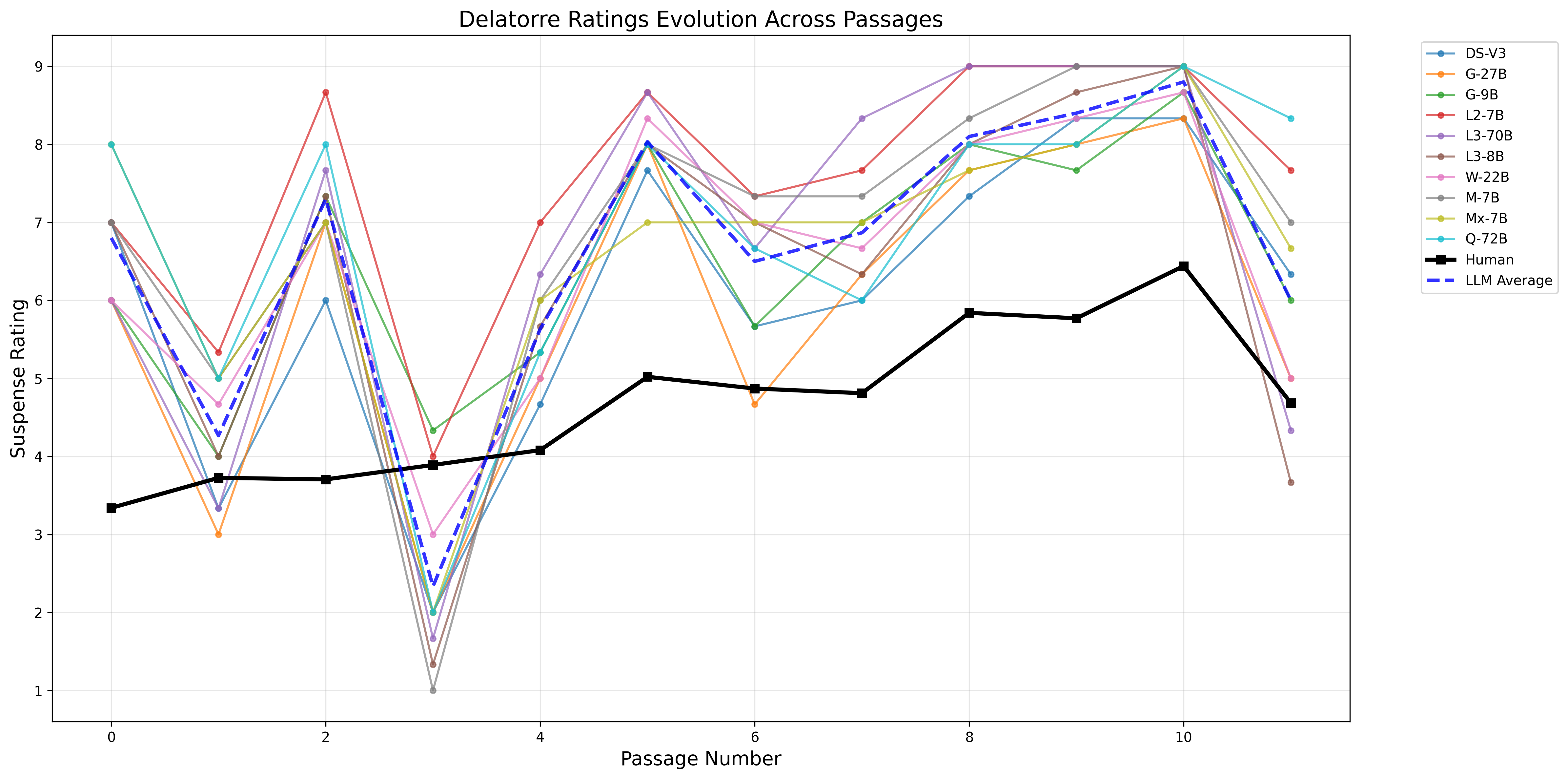}
        \subcaption[.]{Delatorre Visualization of Model Results.}\label{fig:delatorre-value}
    \end{minipage}
    \newline
    \begin{minipage}{1\textwidth}
        \centering
        \includegraphics[width=1\textwidth]{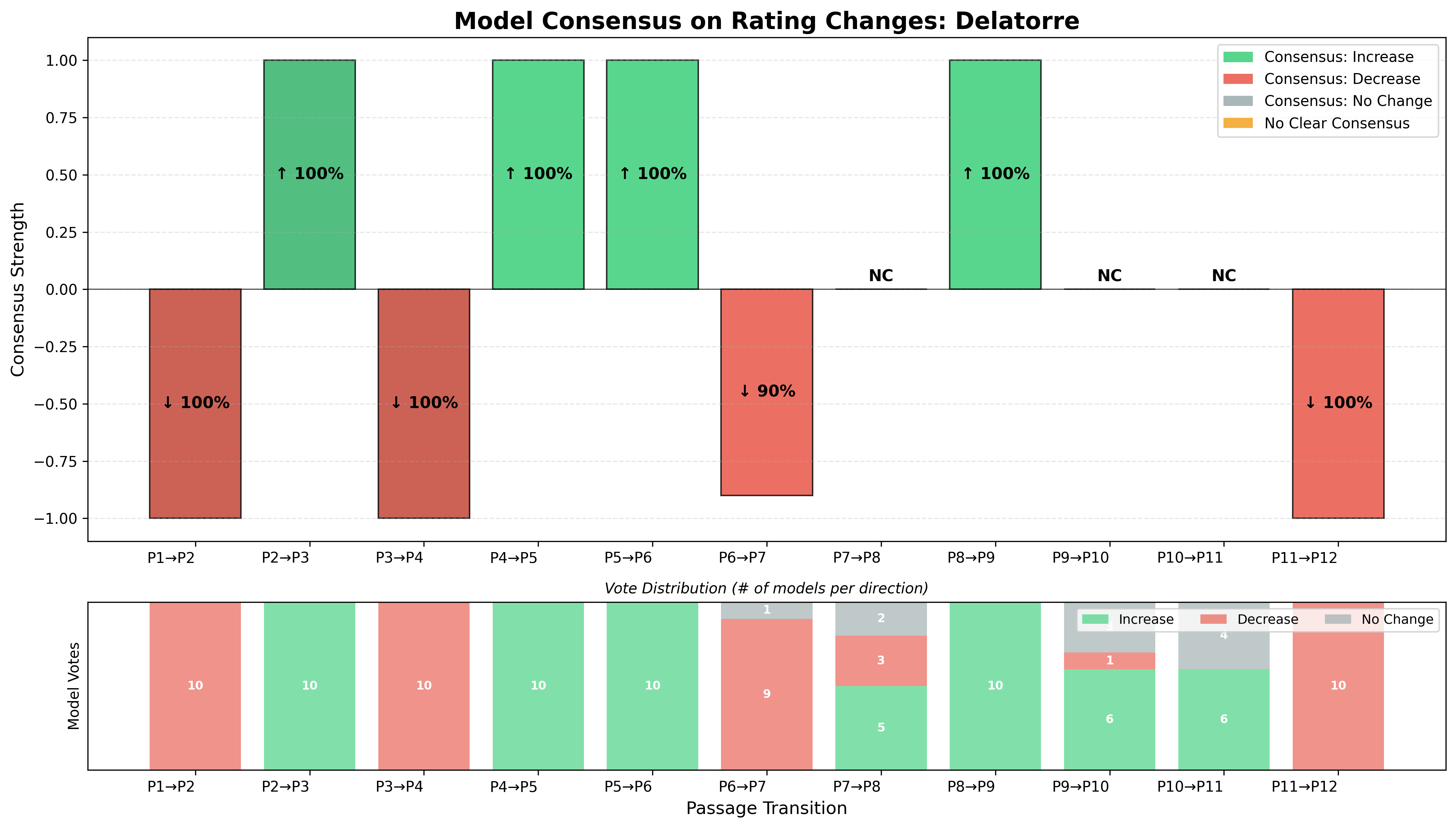}
        \subcaption[.]{Delatorre Visualization of Model Consensus.}\label{fig:delatorre-consensus}
    \end{minipage}
    \label{fig:delatorre-combined}
\end{figure}

\begin{figure}[t]
    \centering
        \begin{minipage}{1\textwidth}
        \centering
        \includegraphics[width=1\textwidth]{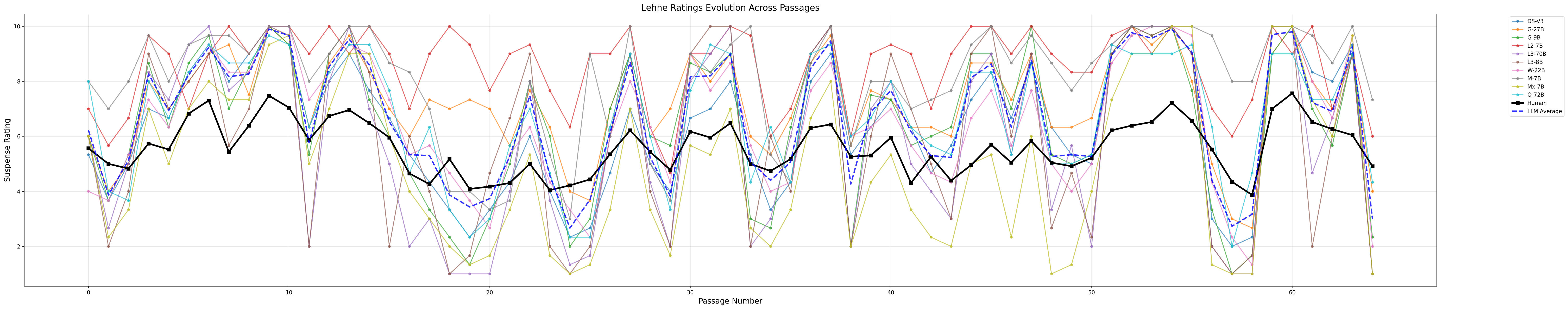}
        \subcaption[fourth caption.]{Lehne  directionality.}\label{fig:lehne-direction}
    \end{minipage}
    \newline
    \begin{minipage}{1\textwidth}
        \centering
        \includegraphics[width=1\textwidth]{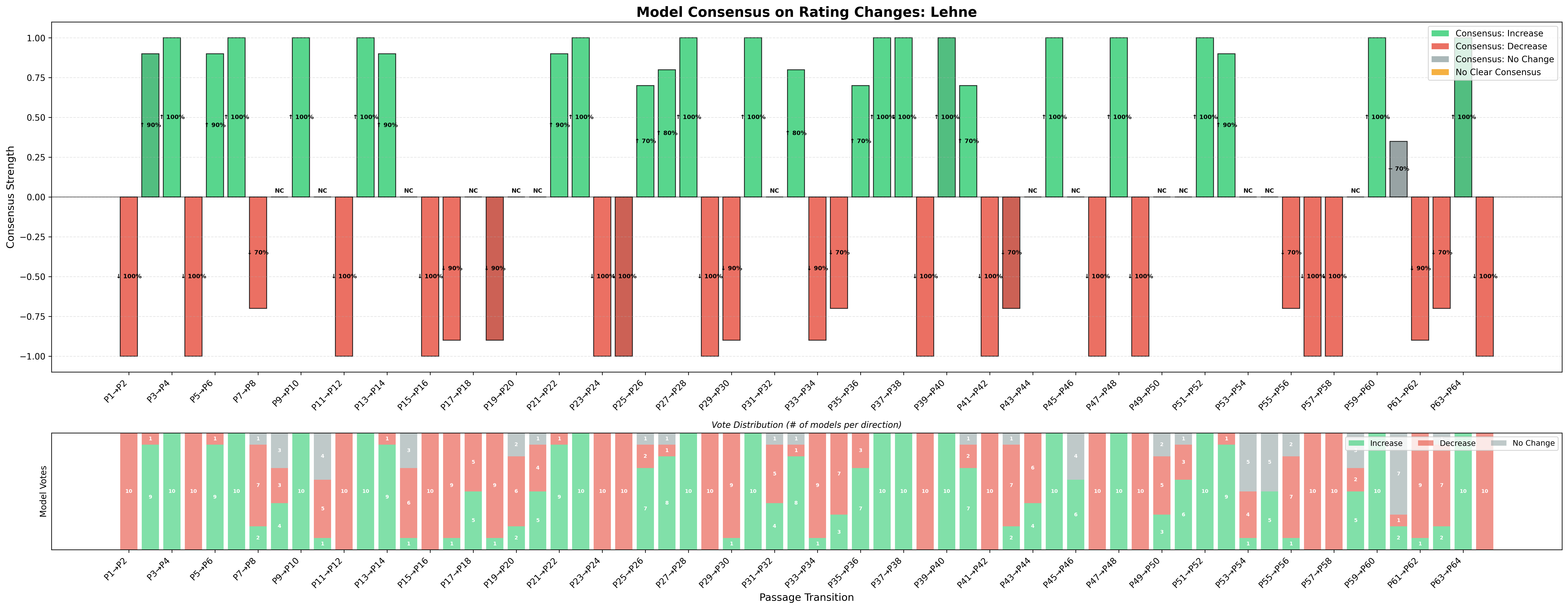}
        \subcaption[fourth caption.]{Lehne inflection points.}\label{fig:lehne-inflection-points}
    \end{minipage}
    \label{fig:lehne-combined}
\end{figure}

\section{Adversarial Story Experiments}
\label{sec:adversarial}

According to psychological models of suspense, complex cognitive processes--such as reasoning about the probability and uncertainty of potential story outcomes--underlie the experience of suspense in humans \citep{Gerrig1994-zo, Brewer1988-va, Delatorre2018-fl, Lehne2015-jw}. Brewer's Structural-Affect Theory explicitly connects suspense to emotional uncertainty resulting from delayed narrative outcomes~\citep{Brewer1988-va}. Lehne's model emphasizes uncertainty and anticipation about the narrative's future outcomes as key drivers of suspense~\citep{Lehne2015-jw}. Similarly, \citet{Delatorre2018-fl} experimentally validated suspense as dependent on outcome uncertainty by demonstrating significant suspense reduction when narrative endings were revealed in advance. Given that suspense fundamentally relies on such complex cognitive mechanisms, the question arises: why should an LM, devoid of human cognitive faculties, be able to detect and rate suspense in narrative texts? Are LMs simply triggering on keywords or have they internalized textual patterns and narrative tropes associated with suspense? Alternatively, might LMs exhibit a form of ``reasoning'' about potential outcomes, as activation patterns propagate through successive layers of their network architectures? Lastly, could their performance reflect exposure to the very texts or types of studies typically used in psychological research?

To investigate how LMs process stories for suspense detection, we conduct 11 types of adversarial permutations on stories from \citet{Gerrig1994-zo} and \citet{Delatorre2018-fl}. These adversarial attacks target surface forms (e.g., word-order manipulations) and shallow semantic levels (e.g., name substitutions), aiming to increase the reader’s cognitive load. Forecasting potential outcomes (a key element of many suspense theories) is already a high-cognitive-load process \citep{Graesser1994}. Increasing demands on cognitive resources has been shown to interfere with inferential comprehension and prediction \citep{Kahneman1973}, suggesting that when readers must devote attentional capacity to resolving text disruptions, fewer resources remain for generating or updating outcome forecasts. Consequently, if suspense depends upon these predictive inferences, deliberately raising the cognitive load could diminish suspense perceptions. Our adversarial permutations are chosen to be implementable at scale without the need for the careful plot treatment designs such as was done in the original Gerrig study. These adversarial tests serve as a stress test for both human readers and LMs to probe whether and how they engage in such forecasting under heightened cognitive demands.

\begin{figure}[htbp]
    \centering
    \includegraphics[width=\linewidth]{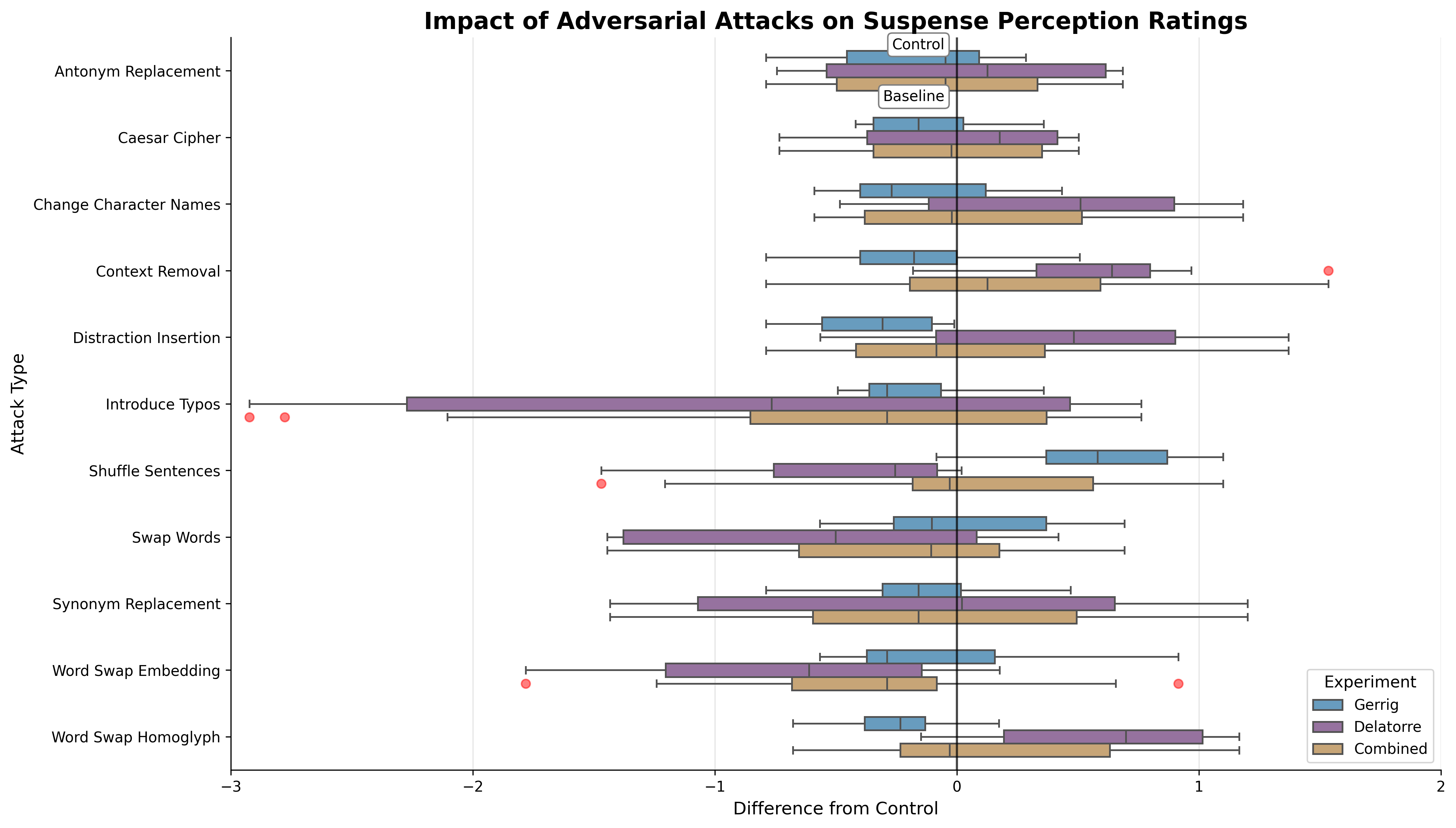}
    \caption{Summary of adversarial results on Gerrig and Delatorre.}
    \label{fig:adversarial-summary}
\end{figure}

Our adversarial permutations are as follows:

\begin{enumerate}

    \item \textbf{Word Swapping}  
    Pairs of words within each passage are swapped. This is destructive to readability from a human perspective. Depending on the number of swaps, the text can become nonsensical, making it difficult for humans to understand. 

    \item \textbf{Antonym Substitution}  
    Available words in each sentence are replaced by their antonyms according to the WordNet lexical database, with probability of 80\%. Each passage is limited to having between 1 and 50 antonyms.

    \item \textbf{Synonym Substitution}  
    Between one and ten words per passage are replaced with synonyms, with the same approach and parameters used for Antonym Substitution.

    \item \textbf{Caesar Ciphering} 
    Text is ciphered using a Caesar cipher with a step size of 3. This method is idealized as being highly disruptive to a human reader. For this attack, the prompt is modified to include information about the cipher and the step size.

    \item \textbf{Homoglyph Substitution}  
    Words are replaced with visually similar homoglyphs. This attack is idealized as being barely perceivable to humans.

    \item \textbf{Distraction Phrase Insertion}  
    A series of preset distraction phrases are inserted according to a "budget" of one distraction every five sentences. This phrase is unrelated to the text but is semantically meaningful. We do not anticipate that the phrase would be factored in the human understanding of suspense, and do not include relevant suspense trigger words.

    \item \textbf{Sentence Shuffling}  
    Sentences are globally shuffled, altering the semantic content of the text.

    \item \textbf{Embedding-Based Word Substitution}  
    Words are swapped based on their nearest neighbors in GloVe \citep{Pennington2014GLoVe} embeddings.

    \item \textbf{Character Name Changes}  
    Character names were changed. 

    \item \textbf{Context Removal}  
    Sentences with sentiment scores below a threshold (0.5 compound sentiment) were removed, measured via VADER \citep{Hutto2014VADERAP} sentiment analysis model. 
    
    \item \textbf{Typo insertion}
    Up to 10 characters per sentence are replaced with random characters.
\end{enumerate}

We run the same experimental setup from \refsec{experimental-setup} but with our permuted story segments.
For each of the adversarial attacks, we expect human ratings of suspense to {\bf drop} because the texts are made unreadable or nonsensical in different ways to increase the cognitive load of reading.
Thus we look for situations where the LM's ratings {\bf remain the same}, or {\bf rise} as evidence that LMs are using very different processes to generate ratings.

\reffig{adversarial-summary} shows the changes in LM ratings when each adversarial permutation is applied across all story segments from \citet{Gerrig1994-zo} and \citet{Delatorre2018-fl}. 
For most attacks, the average LM response remains unchanged, with less than a point of rating difference in either direction. 
Some attacks cause a {\em rise} in suspense in the \cite{Delatorre2018-fl} story segments, especially when character names are changed or distraction sentences are inserted.
Because models should be able to operate on all types of stories, the results on the combined \citet{Gerrig1994-zo} and \citet{Delatorre2018-fl} story segments do not produce any significant change in LM rating behavior.
This suggests that the LMs are using different processes than those believed to be used by humans. 
LMs are able to look past the surface and semantic alterations made by the adversarial attacks.
\section{Conclusions}
\label{sec:conclusion}

In this study, we investigated the 
question of whether language models can rate story suspense similarly to humans.
Following strict adherence to guidelines on {\em machine psychology}~\cite{Hagendorff2023MachinePI}, we replicated the results of four seminal human suspense judgement experiments.
The results are mixed.
LMs generally are not capable of providing numerical ratings of suspense that correspond closely to human average numerical ratings.
However, it may be the case that one only cares about binary judgments of suspense presence or absence, in which case LMs appear quite competent, at least for the limited selection of stories selected for the original psychology studies.
LMs are also much more competent with judgments of rising or falling suspense.
This could be useful in applications that help human writers assess how their audiences will perceive their stories~\cite{oneill:cc2009,Lin2023Ontology,Lin2023Beyond}.
Unfortunately, the points where suspense switches directions from rising to falling, or vice versa, are likely to be the most crucial points in a story, and this is where LMs have the least accuracy in directionality judgments. 

We find evidence that LMs do not process suspense the same way that cognitive theories indicate humans process suspense. 
Adversarial attacks did not have the anticipated effects, and in some cases had the opposite effect.
We also find that LMs do not read stories like humans, building up a model of the story across segments as cognitive models of human suspense perception suggest. 
They appear to focus solely on cues within segments.

\section{Ethics}

We investigate the extent to which Large Language Models (LMs) align with human perception of suspense in textual narratives. All studies that we compare with our models were obtained in accordance with relevant data usage policies. All authors from previous studies and narratives were cited in our work. No human subjects involved in our survey or experiments outside of computational observations.

A potential harm of our work is the overgeneralization of LM capabilities. Although LMs demonstrate proficiency in text-based analysis of suspense, this does not imply that LMs possess human-like understanding of complex psychological phenomena. 

Our studies come from diverse cultural backgrounds—including English, American, Hungarian, Spanish, and German sources; however, our LMs are fed English-translated versions of the original narratives, so subtle linguistic and cultural nuances may be lost. We also want to emphasize that suspense is not universally perceived in the same way. Factors such as age, gender, and socioeconomic status may greatly influence how individuals experience uncertainty and tension in narratives. We wanted to recognize that further research can be done into how these variations can affect how human perception of suspense compares to LMs.

\section*{Acknowledgments} \label{sec:acknowledgements}
The authors collectively would like to thank our anonymous reviewers for their comments and feedback. We appreciate the help provided by Madhura Keshava Ummettuguli with exploring the ideas during the conception of this paper. The costs for some of the model inference in experiments were covered in part using the generous support to Glenn Matlin from \textsc{TogetherAI}.

\bibliography{lab,paperpile,more}

\begin{thebibliography}{56}
\providecommand{\natexlab}[1]{#1}
\providecommand{\url}[1]{\texttt{#1}}
\expandafter\ifx\csname urlstyle\endcsname\relax
  \providecommand{\doi}[1]{doi: #1}\else
  \providecommand{\doi}{doi: \begingroup \urlstyle{rm}\Url}\fi

\bibitem[Bentz et~al.(2024)Bentz, Cortez~Espinoza, Simeonova, K{\"{o}}ppe, and
  Onea]{Bentz2024-qu}
Maria Bentz, Maya Cortez~Espinoza, Vesela Simeonova, Tilmann K{\"{o}}ppe, and
  Edgar Onea.
\newblock Measuring suspense in real time: A new experimental methodology.
\newblock \emph{Sci. Study Lit.}, 12\penalty0 (1):\penalty0 92--112, 18~April
  2024.

\bibitem[Branigan(1992)]{branigan92}
E.~Branigan.
\newblock \emph{Narrative Comprehension and Film}.
\newblock Routledge, New York, 1992.

\bibitem[Brewer(1996)]{Brewer1996-rb}
W~F Brewer.
\newblock The nature narrative suspense problem rereading.
\newblock In P~Vorderer, H~J Wulff, and M~Friedrichsen (eds.), \emph{Suspense:
  Conceptualizations, Theoretical Analyses, Empirical Explorations}, pp.\
  107--127. Lawrence Erlbaum Associates, Mahwah, NJ, 1996.

\bibitem[Brewer \& Lichtenstein(1982)Brewer and Lichtenstein]{Brewer1982-ml}
William~F Brewer and Edward~H Lichtenstein.
\newblock Stories are to entertain: A structural-affect theory of stories.
\newblock \emph{J. Pragmat.}, 6\penalty0 (5-6):\penalty0 473--486, December
  1982.

\bibitem[Brewer \& Ohtsuka(1988)Brewer and Ohtsuka]{Brewer1988-va}
William~F Brewer and Keisuke Ohtsuka.
\newblock Story structure, characterization, just world organization, and
  reader affect in american and hungarian short stories.
\newblock \emph{Poetics (Amst.)}, 17\penalty0 (4-5):\penalty0 395--415, October
  1988.

\bibitem[Carroll(1984)]{Carroll1984-zw}
N~Carroll.
\newblock Toward theory film suspense.
\newblock \emph{Persistence Vision}, 1:\penalty0 65--89, 1984.

\bibitem[Cheong \& Young(2015)Cheong and Young]{Cheong2015SuspenserAS}
Yun-Gyung Cheong and Robert~Michael Young.
\newblock Suspenser: A story generation system for suspense.
\newblock \emph{IEEE Transactions on Computational Intelligence and AI in
  Games}, 7:\penalty0 39--52, 2015.
\newblock URL \url{https://api.semanticscholar.org/CorpusID:9179175}.

\bibitem[Colby et~al.(1989)Colby, Ortony, Clore, and Collins]{Colby1989-ex}
B~N Colby, Andrew Ortony, Gerald~L Clore, and Allan Collins.
\newblock The cognitive structure of emotions.
\newblock \emph{Contemp. Sociol.}, 18\penalty0 (6):\penalty0 957, November
  1989.

\bibitem[Delatorre et~al.(2018)Delatorre, Le\'{o}n, Salguero, Palomo-Duarte,
  and Gerv\'{a}s]{Delatorre2018-fl}
Pablo Delatorre, Carlos Le\'{o}n, Alberto Salguero, Manuel Palomo-Duarte, and
  Pablo Gerv\'{a}s.
\newblock Confronting a paradox: A new perspective of the impact of uncertainty
  in suspense.
\newblock \emph{Front. Psychol.}, 9:\penalty0 1392, 8~August 2018.

\bibitem[Delatorre et~al.(2020)Delatorre, Le{\'o}n, Salguero, and
  Tapscott]{Delatorre2020PredictingTE}
Pablo Delatorre, Carlos Le{\'o}n, Alberto~G. Salguero, and Alan Tapscott.
\newblock Predicting the effects of suspenseful outcome for automatic
  storytelling.
\newblock \emph{Knowl. Based Syst.}, 209:\penalty0 106450, 2020.
\newblock URL \url{https://api.semanticscholar.org/CorpusID:224867428}.

\bibitem[Delatorre et~al.(2021)Delatorre, Le{\'o}n, and
  Hidalgo]{Delatorre2021ImprovingTF}
Pablo Delatorre, Carlos Le{\'o}n, and Alberto~Salguero Hidalgo.
\newblock Improving the fitness function of an evolutionary suspense generator
  through sentiment analysis.
\newblock \emph{IEEE Access}, 9:\penalty0 39626--39635, 2021.
\newblock URL \url{https://api.semanticscholar.org/CorpusID:232317642}.

\bibitem[Doust \& Piwek(2017)Doust and Piwek]{Doust2017-ut}
Richard Doust and Paul Piwek.
\newblock A model of suspense for narrative generation.
\newblock In \emph{Proceedings of the 10th International Conference on Natural
  Language Generation}, Stroudsburg, PA, USA, 2017. Association for
  Computational Linguistics.

\bibitem[Fendt \& Young(2017)Fendt and Young]{Fendt2017LeveragingIR}
Matthew~William Fendt and Robert~Michael Young.
\newblock Leveraging intention revision in narrative planning to create
  suspenseful stories.
\newblock \emph{IEEE Transactions on Computational Intelligence and AI in
  Games}, 9:\penalty0 381--392, 2017.
\newblock URL \url{https://api.semanticscholar.org/CorpusID:38348125}.

\bibitem[Freytag(1894)]{Freytag1894-zz}
G~Freytag.
\newblock \emph{Freytag, {G}. (1894). Freytag's Technique Drama: Exposition
  Dramatic Composition Art}.
\newblock Scott Foresman, Chicago, IL, 1894.

\bibitem[Gerrig(1994)]{Gerrig1994-et}
Richard Gerrig.
\newblock Experiencing narrative worlds: on the psychological activities of
  reading.
\newblock \emph{Choice (Middletown)}, 31\penalty0 (07):\penalty0
  31--3591--31--3591, 1~March 1994.

\bibitem[Gerrig \& Bernardo(1994)Gerrig and Bernardo]{Gerrig1994-zo}
Richard~J Gerrig and Allan B~I Bernardo.
\newblock Readers as problem-solvers in the experience of suspense.
\newblock \emph{Poetics (Amst.)}, 22\penalty0 (6):\penalty0 459--472, December
  1994.

\bibitem[Gooding et~al.(2025)Gooding, Lopez-Rivilla, and
  Grefenstette]{Gooding2025-tp}
Sian Gooding, Lucia Lopez-Rivilla, and Edward Grefenstette.
\newblock Writing as a testbed for open ended agents.
\newblock \emph{arXiv [cs.CL]}, 25~March 2025.

\bibitem[Graesser et~al.(1994)Graesser, Singer, and Trabasso]{Graesser1994}
Arthur~C. Graesser, Murray Singer, and Tom Trabasso.
\newblock Constructing inferences during narrative text comprehension.
\newblock \emph{Psychological review}, 101 3:\penalty0 371--95, 1994.
\newblock URL \url{https://api.semanticscholar.org/CorpusID:523868}.

\bibitem[Guidry(2005)]{Guidry2005-wq}
J~Guidry.
\newblock The experience of . . . suspense: understanding the construct, its
  antecedents, and its consequences in consumption and acquisition contexts.
\newblock \emph{Unpublished manuscript}, 2005.

\bibitem[Hagendorff(2023)]{Hagendorff2023MachinePI}
Thilo Hagendorff.
\newblock Machine psychology: Investigating emergent capabilities and behavior
  in large language models using psychological methods.
\newblock \emph{ArXiv}, abs/2303.13988, 2023.
\newblock URL \url{https://api.semanticscholar.org/CorpusID:257757370}.

\bibitem[Haider et~al.(2020)Haider, Eger, Kim, Klinger, and
  Menninghaus]{Haider2020-tr}
Thomas Haider, Steffen Eger, Evgeny Kim, Roman Klinger, and Winfried
  Menninghaus.
\newblock {PO}-{EMO}: Conceptualization, annotation, and modeling of aesthetic
  emotions in german and english poetry.
\newblock \emph{arXiv [cs.CL]}, 17~March 2020.

\bibitem[Hoeken \& van Vliet(2000)Hoeken and van Vliet]{Hoeken2000-jp}
Hans Hoeken and Mario van Vliet.
\newblock Suspense, curiosity, and surprise: How discourse structure influences
  the affective and cognitive processing of a story.
\newblock \emph{Poetics (Amst.)}, 27\penalty0 (4):\penalty0 277--286, May 2000.

\bibitem[Hsieh(2021)]{Hsieh2021-ef}
Chihmao Hsieh.
\newblock Suspense: exploring a new lens for outcome uncertainty in
  entrepreneurship.
\newblock \emph{Int. J. Entrep. Ventur.}, 13\penalty0 (1):\penalty0 27, 2021.

\bibitem[Hutto \& Gilbert(2014)Hutto and Gilbert]{Hutto2014VADERAP}
Clayton~J. Hutto and Eric Gilbert.
\newblock Vader: A parsimonious rule-based model for sentiment analysis of
  social media text.
\newblock \emph{Proceedings of the International AAAI Conference on Web and
  Social Media}, 2014.
\newblock URL \url{https://api.semanticscholar.org/CorpusID:12233345}.

\bibitem[Kahneman(1973)]{Kahneman1973}
Daniel Kahneman.
\newblock \emph{Attention and Effort}.
\newblock Prentice-Hall, Englewood Cliffs, NJ, 1973.

\bibitem[Kaspar et~al.(2016)Kaspar, Zimmermann, and Wilbers]{Kaspar2016-bl}
Kai Kaspar, Daniel Zimmermann, and Anne-Kathrin Wilbers.
\newblock Thrilling news revisited: The role of suspense for the enjoyment of
  news stories.
\newblock \emph{Front. Psychol.}, 7:\penalty0 1913, 15~December 2016.

\bibitem[Lehne \& Koelsch(2015)Lehne and Koelsch]{Lehne2015-jw}
Moritz Lehne and Stefan Koelsch.
\newblock Toward a general psychological model of tension and suspense.
\newblock \emph{Front. Psychol.}, 6, 11~February 2015.

\bibitem[Li et~al.(2021)Li, Bramley, and Gureckis]{Li2021-di}
Zhi-Wei Li, Neil~R Bramley, and Todd~M Gureckis.
\newblock Expectations about future learning influence moment-to-moment
  feelings of suspense.
\newblock \emph{Cogn. Emot.}, 35\penalty0 (6):\penalty0 1099--1120, 18~August
  2021.

\bibitem[Liang et~al.(2022)Liang, Bommasani, Lee, Tsipras, Soylu, Yasunaga,
  Zhang, Narayanan, Wu, Kumar, Newman, Yuan, Yan, Zhang, Cosgrove, Manning,
  R\'{e}, Acosta-Navas, Hudson, Zelikman, Durmus, Ladhak, Rong, Ren, Yao, Wang,
  Santhanam, Orr, Zheng, Yuksekgonul, Suzgun, Kim, Guha, Chatterji, Khattab,
  Henderson, Huang, Chi, Xie, Santurkar, Ganguli, Hashimoto, Icard, Zhang,
  Chaudhary, Wang, Li, Mai, Zhang, and Koreeda]{Liang2022-ew}
Percy Liang, Rishi Bommasani, Tony Lee, Dimitris Tsipras, Dilara Soylu,
  Michihiro Yasunaga, Yian Zhang, Deepak Narayanan, Yuhuai Wu, Ananya Kumar,
  Benjamin Newman, Binhang Yuan, Bobby Yan, Ce~Zhang, Christian Cosgrove,
  Christopher~D Manning, Christopher R\'{e}, Diana Acosta-Navas, Drew~A Hudson,
  Eric Zelikman, Esin Durmus, Faisal Ladhak, Frieda Rong, Hongyu Ren, Huaxiu
  Yao, Jue Wang, Keshav Santhanam, Laurel Orr, Lucia Zheng, Mert Yuksekgonul,
  Mirac Suzgun, Nathan Kim, Neel Guha, Niladri Chatterji, Omar Khattab, Peter
  Henderson, Qian Huang, Ryan Chi, Sang~Michael Xie, Shibani Santurkar, Surya
  Ganguli, Tatsunori Hashimoto, Thomas Icard, Tianyi Zhang, Vishrav Chaudhary,
  William Wang, Xuechen Li, Yifan Mai, Yuhui Zhang, and Yuta Koreeda.
\newblock Holistic evaluation of language models.
\newblock \emph{arXiv [cs.CL]}, 16~November 2022.

\bibitem[Lin \& Riedl(2023)Lin and Riedl]{Lin2023Ontology}
Zhiyu Lin and M.~O. Riedl.
\newblock An ontology of co-creative ai systems.
\newblock In \emph{Proceedings of the 2024 NeurIPS Workshop on Machine Learning
  for Creativity and Design}, 2023.
\newblock URL \url{https://arxiv.org/abs/2310.07472}.

\bibitem[Lin et~al.(2023)Lin, Ehsan, Agarwal, Dani, Vashishth, and
  Riedl]{Lin2023Beyond}
Zhiyu Lin, Upol Ehsan, Rohan Agarwal, Samihan Dani, Vidushi Vashishth, and
  Mark~O. Riedl.
\newblock Beyond prompts: Exploring the design space of mixed-initiative
  co-creativity systems.
\newblock In \emph{Proceedings of the 2023 International Conference on
  Computational Creativity}, 2023.
\newblock URL \url{https://arxiv.org/abs/2305.07465}.

\bibitem[Liu(2012)]{Liu2012-sy}
Bing Liu.
\newblock \emph{Sentiment Analysis and Opinion Mining}.
\newblock Springer International Publishing, Cham, 2012.

\bibitem[L{\"{o}}hn et~al.(2024)L{\"{o}}hn, Kiehne, Ljapunov, and
  Balke]{Lohn2024-jg}
Lea L{\"{o}}hn, Niklas Kiehne, Alexander Ljapunov, and Wolf-Tilo Balke.
\newblock Is machine psychology here? on requirements for using human
  psychological tests on large language models.
\newblock \emph{Int Conf Nat Lang Gener}, pp.\  230--242, 2024.

\bibitem[Madrigal et~al.(2011)Madrigal, Bee, Chen, and
  LaBarge]{Madrigal2011-dx}
Robert Madrigal, Colleen Bee, Johnny Chen, and Monica LaBarge.
\newblock The effect of suspense on enjoyment following a desirable outcome:
  The mediating role of relief.
\newblock \emph{Media Psychol.}, 14\penalty0 (3):\penalty0 259--288, 31~August
  2011.

\bibitem[Malberg et~al.(2024)Malberg, Poletukhin, Schuster, and
  Groh]{Malberg2024-bz}
Simon Malberg, Roman Poletukhin, Carolin~M Schuster, and Georg Groh.
\newblock A comprehensive evaluation of cognitive biases in {LLMs}.
\newblock \emph{arXiv [cs.CL]}, 20~October 2024.

\bibitem[Moulard et~al.(2012)Moulard, Kroff, and Folse]{Moulard2012-pi}
Julie~Guidry Moulard, Michael~W Kroff, and Judith Anne~Garretson Folse.
\newblock Unraveling consumer suspense: The role of hope, fear, and probability
  fluctuations.
\newblock \emph{J. Bus. Res.}, 65\penalty0 (3):\penalty0 340--346, March 2012.

\bibitem[O'Neill \& Riedl(2011)O'Neill and Riedl]{oneill:acii2011}
Brian O'Neill and Mark Riedl.
\newblock Toward a computational framework of suspense and dramatic arc.
\newblock In \emph{Proceedings of the 4th International Conference on Affective
  Computing and Intelligent Interaction}, pp.\  246--255, Memphis, Tennessee,
  October 2011.
\newblock URL \url{http://www.cc.gatech.edu/~riedl/pubs/acii11.pdf}.

\bibitem[O'Neill \& Riedl(2014)O'Neill and Riedl]{O-Neill2014-ta}
Brian O'Neill and Mark Riedl.
\newblock Dramatis: A computational model of suspense.
\newblock \emph{Proc. Conf. AAAI Artif. Intell.}, 28\penalty0 (1), 21~June
  2014.

\bibitem[O'Neill \& Riedl(2009)O'Neill and Riedl]{oneill:cc2009}
Brian O'Neill and Mark~O. Riedl.
\newblock Supporting human creative story authoring with a synthetic audience.
\newblock In \emph{Proceedings of the 7th Creativity and Cognition Conference},
  Berkeley, California, October 2009.
\newblock URL \url{http://www.cc.gatech.edu/~riedl/pubs/oneill-cc09.pdf}.

\bibitem[Paech(2023)]{Paech2023-bt}
Samuel~J Paech.
\newblock Creative writing bench.
\newblock \url{https://eqbench.com/creative\_writing.html}, 2023.
\newblock Accessed: 2025-3-27.

\bibitem[Palich \& Bagby(1995)Palich and Bagby]{Palich1995-ao}
L~E Palich and D~R Bagby.
\newblock Using cognitive theory explain entrepreneurial risk-taking -
  challenging conventional wisdom.
\newblock \emph{Journal Business Venturing}, 10:\penalty0 425--438, 1995.

\bibitem[Pennington et~al.(2014)Pennington, Socher, and
  Manning]{Pennington2014GLoVe}
Jeffrey Pennington, Richard Socher, and Christopher Manning.
\newblock {G}lo{V}e: Global vectors for word representation.
\newblock In Alessandro Moschitti, Bo~Pang, and Walter Daelemans (eds.),
  \emph{Proceedings of the 2014 Conference on Empirical Methods in Natural
  Language Processing ({EMNLP})}, pp.\  1532--1543, Doha, Qatar, October 2014.
  Association for Computational Linguistics.
\newblock \doi{10.3115/v1/D14-1162}.
\newblock URL \url{https://aclanthology.org/D14-1162/}.

\bibitem[Peterson \& Raney(2008)Peterson and Raney]{Peterson2008-tp}
Erik~M Peterson and Arthur~A Raney.
\newblock Reconceptualizing and reexamining suspense as a predictor of mediated
  sports enjoyment.
\newblock \emph{J. Broadcast. Electron. Media}, 52\penalty0 (4):\penalty0
  544--562, 7~November 2008.

\bibitem[Piper et~al.(2021)Piper, So, and Bamman]{Piper2021-ql}
Andrew Piper, Richard~Jean So, and David Bamman.
\newblock Narrative theory for computational narrative understanding.
\newblock In Marie-Francine Moens, Xuanjing Huang, Lucia Specia, and Scott
  Wen-Tau Yih (eds.), \emph{Proceedings of the 2021 Conference on Empirical
  Methods in Natural Language Processing}, pp.\  298--311, Stroudsburg, PA,
  USA, November 2021. Association for Computational Linguistics.

\bibitem[Plaza-del Arco et~al.(2024)Plaza-del Arco, Curry, Curry, and
  Hovy]{Plaza-del-Arco2024-kx}
Flor~Miriam Plaza-del Arco, Alba Curry, Amanda~Cercas Curry, and Dirk Hovy.
\newblock Emotion analysis in {NLP}: Trends, gaps and roadmap for future
  directions.
\newblock \emph{arXiv [cs.CL]}, 2~March 2024.

\bibitem[Shah et~al.(2023)Shah, Marupudi, Koenen, Bhardwaj, and
  Varma]{Shah2023-ej}
Raj Shah, Vijay Marupudi, Reba Koenen, Khushi Bhardwaj, and Sashank Varma.
\newblock Numeric magnitude comparison effects in large language models.
\newblock In Anna Rogers, Jordan Boyd-Graber, and Naoaki Okazaki (eds.),
  \emph{Findings of the Association for Computational Linguistics: ACL 2023},
  pp.\  6147--6161, Toronto, Canada, July 2023. Association for Computational
  Linguistics.

\bibitem[Shaikh et~al.(2024)Shaikh, Dandekar, Panat, and
  Dandekar]{Shaikh2024-bv}
Ammar Shaikh, Raj~Abhijit Dandekar, Sreedath Panat, and Rajat Dandekar.
\newblock {CBEval}: A framework for evaluating and interpreting cognitive
  biases in {LLMs}.
\newblock \emph{arXiv [cs.CL]}, 4~December 2024.

\bibitem[Shaki et~al.(2023)Shaki, Kraus, and Wooldridge]{Shaki2023-eg}
Jonathan Shaki, Sarit Kraus, and Michael Wooldridge.
\newblock Cognitive effects in large language models.
\newblock \emph{arXiv [cs.AI]}, 28~August 2023.

\bibitem[Steg et~al.(2022)Steg, Slot, and Pianzola]{Steg2022-lx}
Max Steg, Karlo H~R Slot, and Federico Pianzola.
\newblock Computational detection of narrativity: A comparison using textual
  features and reader response.
\newblock \emph{Digital Humanities Quarterly}, 2022.

\bibitem[Vaswani et~al.(2017)Vaswani, Shazeer, Parmar, and
  {others}]{Vaswani2017-qo}
A~Vaswani, N~Shazeer, N~Parmar, and {others}.
\newblock Attention is all you need.
\newblock \emph{Adv. Neural Inf. Process. Syst.}, 2017.

\bibitem[Vorderer et~al.(1996)Vorderer, Wulff, and
  Friedrichsen]{Vorderer1996-xj}
Peter Vorderer, Hans~J{\"{u}}rgen Wulff, and Mike Friedrichsen.
\newblock \emph{Suspense: Conceptualizations, theoretical analyses, and
  empirical explorations}.
\newblock Routledge, London, England, 1996.

\bibitem[Wang et~al.(2024)Wang, Ma, Zhang, Ni, Chandra, Guo, Ren, Arulraj, He,
  Jiang, Li, Ku, Wang, Zhuang, Fan, Yue, and Chen]{Wang2024-pk}
Yubo Wang, Xueguang Ma, Ge~Zhang, Yuansheng Ni, Abhranil Chandra, Shiguang Guo,
  Weiming Ren, Aaran Arulraj, Xuan He, Ziyan Jiang, Tianle Li, Max Ku, Kai
  Wang, Alex Zhuang, Rongqi Fan, Xiang Yue, and Wenhu Chen.
\newblock {MMLU}-pro: A more robust and challenging multi-task language
  understanding benchmark.
\newblock \emph{arXiv [cs.CL]}, 3~June 2024.

\bibitem[Wankhade et~al.(2022)Wankhade, Rao, and Kulkarni]{Wankhade2022-oi}
Mayur Wankhade, Annavarapu Chandra~Sekhara Rao, and Chaitanya Kulkarni.
\newblock A survey on sentiment analysis methods, applications, and challenges.
\newblock \emph{Artif. Intell. Rev.}, 55\penalty0 (7):\penalty0 5731--5780,
  October 2022.

\bibitem[Wilmot \& Keller(2020)Wilmot and Keller]{Wilmot2020-qa}
David Wilmot and Frank Keller.
\newblock Modelling suspense in short stories as uncertainty reduction over
  neural representation.
\newblock In \emph{Proceedings of the 58th Annual Meeting of the Association
  for Computational Linguistics}, Stroudsburg, PA, USA, 2020. Association for
  Computational Linguistics.

\bibitem[Xie \& Riedl(2024)Xie and Riedl]{Xie2024-qv}
Kaige Xie and Mark Riedl.
\newblock Creating suspenseful stories: Iterative planning with large language
  models.
\newblock \emph{arXiv [cs.CL]}, 27~February 2024.

\bibitem[Zhu et~al.(2011)Zhu, Onta\~{n}\'{o}n, and Lewter]{Zhu2011-xq}
Jichen Zhu, Santiago Onta\~{n}\'{o}n, and Brad Lewter.
\newblock Representing game characters' inner worlds through narrative
  perspectives.
\newblock In \emph{Proceedings of the 6th International Conference on
  Foundations of Digital Games}, New York, NY, USA, 29~June 2011. ACM.

\end{thebibliography}
\bibliographystyle{colm2025_conference}

\appendix
\section{Suspense}
\label{app:suspense}

\begin{figure}[htbp]
  \centering
  \begin{minipage}[b]{0.45\linewidth}
    \centering
    \includegraphics[width=\linewidth]{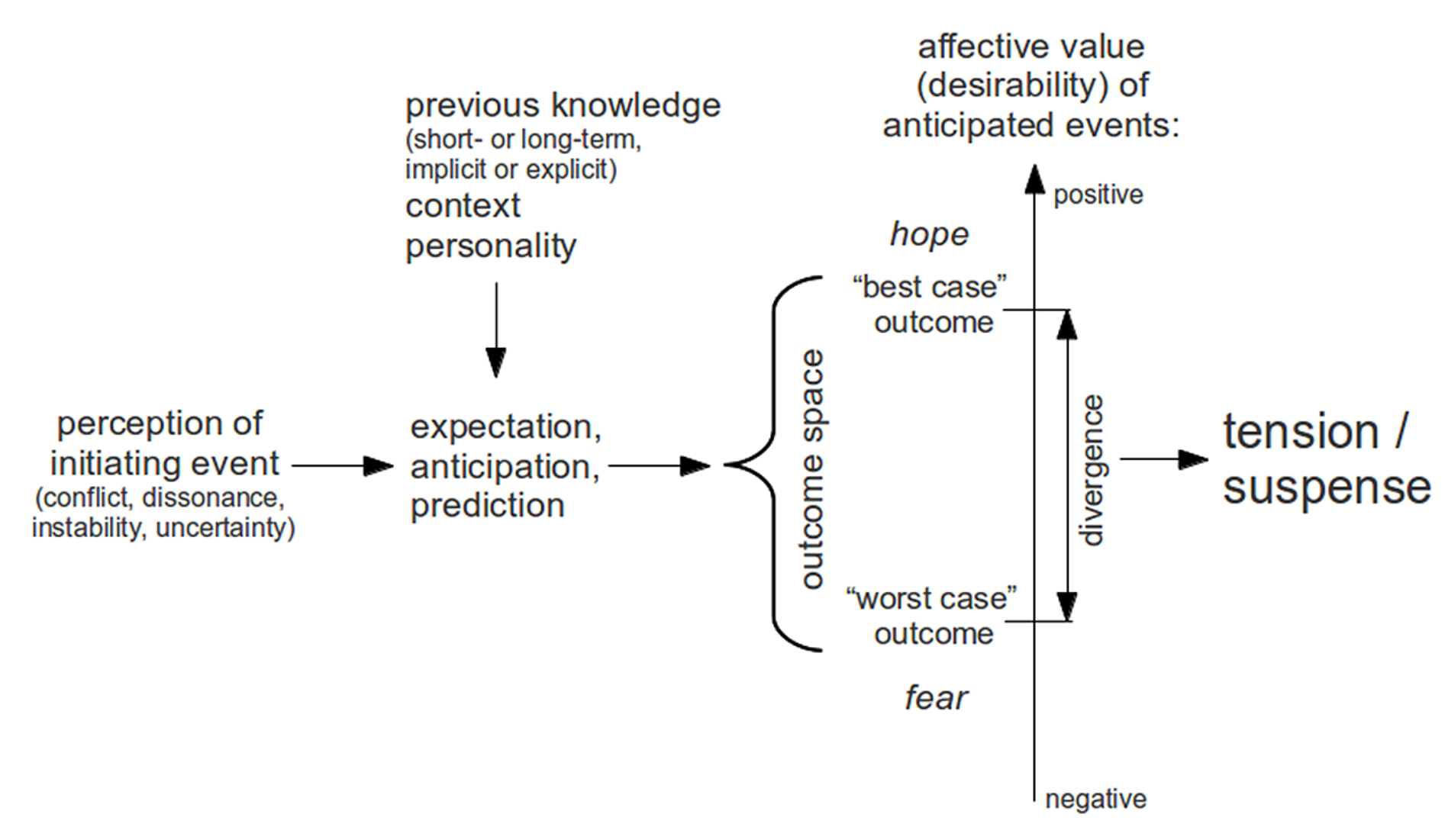}
    \caption{Example of one model of suspense from \citet{Lehne2015-jw} (Figure 2 ``Tension Model'', pg. 7)}
    \label{fig:lehne_suspense_model}
  \end{minipage}\hfill
  \begin{minipage}[b]{0.45\linewidth}
    \centering
    \includegraphics[width=\linewidth]{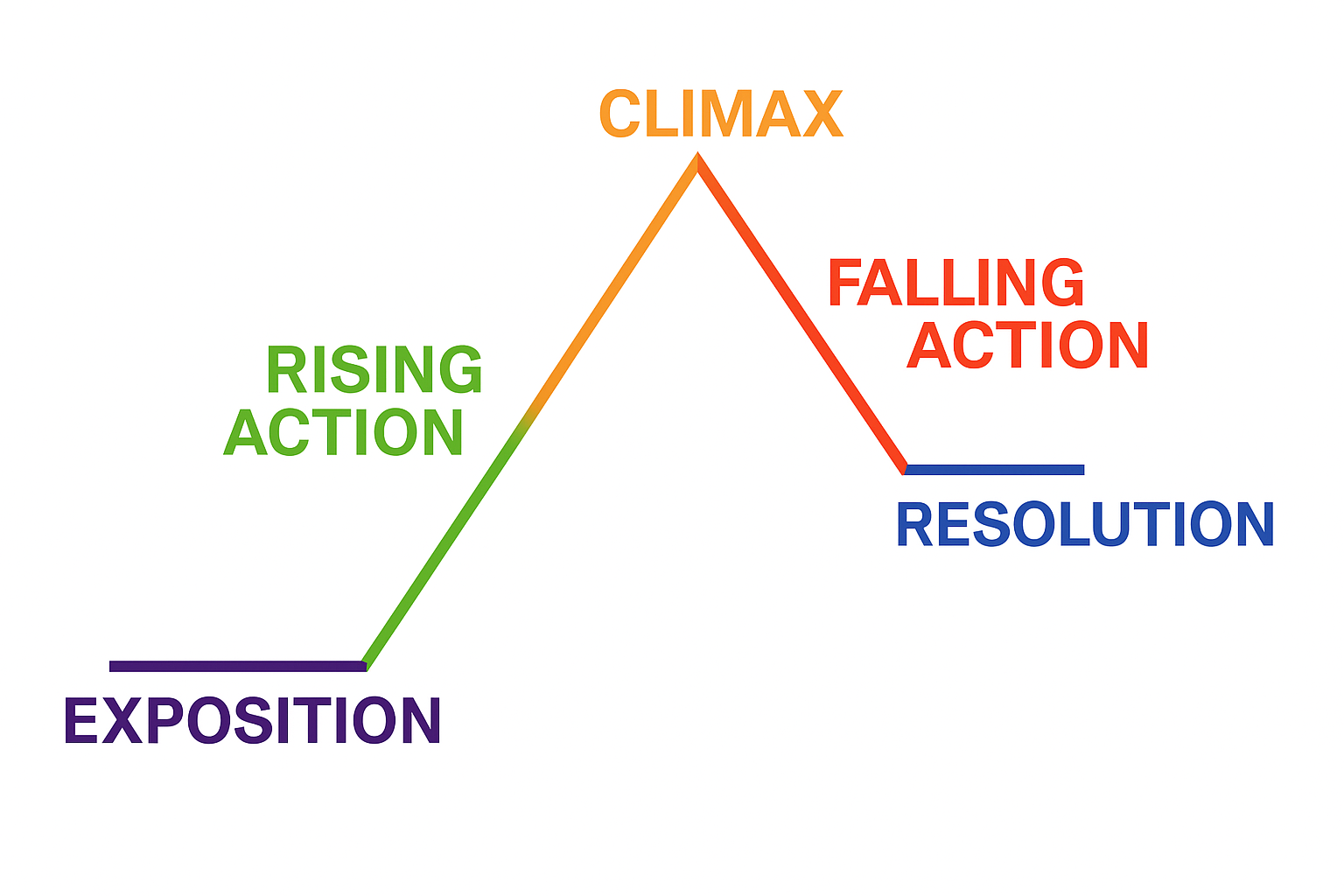}
    \caption{Freytag's Pyramid \citep{Freytag1894-zz}}
    \label{fig:freytags_pyramid}
  \end{minipage}
\end{figure}

\section{Adversarial Attacks}
\label{app:adversarial}

Starting from the datasets and human-annotated labels of \citet{Gerrig1994-zo} and \citet{Delatorre2018-fl}, we employ 11 different adversarial strategies for text attack. Each model-dataset-attack combination is performed three times. Unless explicitly noted, the suspense scores are averaged over the three generations either as a number or a label.
Attacks are performed on a per-passage basis, where a passage is any snippet of text that is followed by a request for a suspense rating. For the \citet{Gerrig1994-zo} dataset, the entire passage and question are shown in a single sequence. \citet{Delatorre2018-fl}, on the other hand, is comprised of 12 different segments, which are each attacked individually.
The following excerpts are examples of each text attack. As a consistent example, we use the first paragraph from the stories from \citet{Delatorre2018-fl} showing the resulting text after each attack method is applied.

\subsection*{Control}
\texttt{
"1 What you are about to read is the story of a real event that took place at the UCSF Benioff Children's Hospital, in San Francisco, California, on 24 February 2008. On that day, since 8 a.m., an eight year old boy called Robert Bent and the entire medical team treating him were all ready for his imminent liver transplant. Just the day before a suitable donor had been found, and they were now awaiting the arrival of the organ. However, they were not sure if Robert would survive the wait as his situation was critical. This is the story of what happened."
}

\subsection*{Word Swapping}
\texttt{
"1 Francisco, that What event in 24 story took read of February at San about is place Benioff you the California, to Children's UCSF Hospital, are a on real the 2008. On and old since 8 year a.m., team Robert for an all eight Bent day, the entire that called transplant. medical boy him ready imminent his treating were liver they donor of found, before the the the had Just day awaiting and a organ. arrival suitable were now been they were Robert his survive sure situation was the critical. would as However, not wait if happened. story is This what the of"
}
\subsection*{Antonym Substitution}
\texttt{
"1 What you differ about to read differ the story of a nominal event that give place at the UCSF Benioff Children's Hospital, in San Francisco, California, on 24 February 2008. On that day, since 8 a. m. , an eight year young boy called Robert Bent and the entire surgical team treating him differ all unready for his imminent liver transplant. Just the day before a suitable donor lack differ lose, and they differ now awaiting the arrival of the organ. However, they differ not uncertain if Robert would succumb the wait as his situation differ noncritical. This differ the story of what dematerialize."
}
\subsection*{Synonym Substitution}
\texttt{
"1 What you live astir to study be the account of a substantial event that took plaza at the UCSF Benioff Nestling 's Infirmary, in San Francisco, Calif., on 24 Feb 2008. On that day, since 8 a. m. , an ashcan school class old boy foretell Robert Bent and the entire medical squad treating him were all quick for his imminent liver graft. Just the day before a suited donor had equal found, and they be at present awaiting the arrival of the electronic organ. However, they were non trusted if Henry m. robert would live on the wait as his berth comprise decisive. This is the write up of what happen."}

\subsection*{Caesar Ciphering}
\texttt{
"1 Zkdw brx duh derxw wr uhdg lv wkh vwrub ri d uhdo hyhqw wkdw wrrn sodfh dw wkh XFVI Ehqlrii Fkloguhq'v Krvslwdo, lq Vdq Iudqflvfr, Fdoliruqld, rq 57 Iheuxdub 5331. Rq wkdw gdb, vlqfh 1 d.p., dq hljkw bhdu rog erb fdoohg Urehuw Ehqw dqg wkh hqwluh phglfdo whdp wuhdwlqj klp zhuh doo uhdgb iru klv lpplqhqw olyhu wudqvsodqw. Mxvw wkh gdb ehiruh d vxlwdeoh grqru kdg ehhq irxqg, dqg wkhb zhuh qrz dzdlwlqj wkh duulydo ri wkh rujdq. Krzhyhu, wkhb zhuh qrw vxuh li Urehuw zrxog vxuylyh wkh zdlw dv klv vlwxdwlrq zdv fulwlfdo. Wklv lv wkh vwrub ri zkdw kdsshqhg.
"
}

\subsection*{Homoglyph Substitution}

\texttt{
"1 What \textbf{y}ou are abo\textbf{u}t \textbf{t}o \textbf{re}ad is the story of a real event tha\textbf{t} took p\textbf{l}a\textbf{c}e \textbf{a}t \textbf{the} UCSF Be\textbf{n}ioff Children\u2019\textbf{s} Hospital, in San Franc\textbf{i}s\textbf{c}o, California, on 24 Febru\textbf{a}ry 200\textbf{8}. On th\textbf{a}t da\textbf{y}, s\textbf{in}ce \textbf{8} a.m., an eight ye\textbf{a}r old boy called Rob\textbf{er}t B\textbf{e}nt and the entire medical team treati\textbf{n}g \textbf{h}im were a\textbf{l}l ready for his immi\textbf{n}ent \textbf{l}iver t\textbf{r}anspla\textbf{n}t. Just the da\textbf{y} before a suitable donor had b\textbf{e}en found, and the\textbf{y} were now awa\textbf{i}ting th\textbf{e} arriva\textbf{l} of the organ. Howe\textbf{ver}, they were not sure if Robert w\textbf{o}ul\textbf{d} \textbf{s}urvive the wait as his situatio\textbf{n} w\textbf{as} crit\textbf{i}c\textbf{a}l. T\textbf{h}is is the story of what happened."
}

\subsection*{Distraction Phrase Insertion}
\texttt{
"1 What you are about to read is the story of a real event that took place at the UCSF Benioff Children's Hospital, in San Francisco, California, on 24 February 2008.  On that day, since 8 a. m. , an eight year old boy called Robert Bent and the entire medical team treating him were all ready for his imminent liver transplant. \textit{He looked for his hidden watch. He couldn't find it}. Just the day before a suitable donor had been found, and they were now awaiting the arrival of the organ.  However, they were not sure if Robert would survive the wait as his situation was critical.  This is the story of what happened. ",
}

\subsection*{Sentence Shuffling}
\texttt{
"1 The analysis showed that it had withstood the impact and it was possible to use the organ for the transplant. When they opened the case, they discovered that the interior bag had ruptured. The two men transporting the liver left the roof via the doorway to the service stairwell, which they decided to walk down. The doctors arrived promptly. However, they were not sure if Robert would survive the wait as his situation was critical."
}

\subsection*{Embedding-Based Word Substitution}
\texttt{
"1 Whereof you constituted roughly dans read is du histories des other reales phenomena that picked site into the UCSF Benioff Children'r Clinic, in Santo Stefania, California, orn 24 February 2008. Orn somebody daytime, because 8 a.m., an seis annual antigua guy titled Robert Bent and per whole medical squad treating he were all ready in his imminent renal transplanted. Virtuous de day beforehand latest suitable donor has played uncovered, ja would were presently awaiting by arriving de nova agency. However, would was nope sure if Richards would live the awaits como her circumstances es crucial. This is de story of what transpired."
}

\subsection*{Character Name Changes}
\texttt{
"1 What you are about to read is the story of a real event that took place at the UCSF Benioff Children's Hospital, in San Francisco, California, on 24 February 2008. On that day, since 8 a.m., an eight year old boy called Alex and the entire medical team treating him were all ready for his imminent liver transplant. Just the day before a suitable donor had been found, and they were now awaiting the arrival of the organ. However, they were not sure if Alex would survive the wait as his situation was critical. This is the story of what happened.
"
}

\subsection*{Context Removal}
\texttt{
"1 What you are about to read is the story of a real event that took place at the UCSF Benioff Children's Hospital, in San Francisco, California, on 24 February 2008. On that day, since 8 a.m., an eight year old boy called Robert Bent and the entire medical team treating him were all ready for his imminent liver transplant. Just the day before a suitable donor had been found, and they were now awaiting the arrival of the organ. This is the story of what happened.
"
}

\subsection*{Typo Insertion}
\texttt{
"1 What you are AF8uF to read is the story of a real event rtWY HK9M 0kZxe at the UCSF Benioff Children's Hospital, in San Francisco, XakkT8rBlW, on 24 e4fFHzEy 2008. On 6uqG day, since 8 a. m. , an eight year old boy called Robert Bent and the entire medical team treating him were all ready for his imminent liver transplant. Just the day before a suitable donor had been found, and they A2tR now awaiting the arrival of the organ. However, they were not dHfW if Robert would survive the wait as his situation was critical. GNJa is the story of what happened.
"
}
\section{Results In Detail}
\label{app:results}

This Appendix gives full-size charts and additional tables for the main results in Section~\refsec{results}.

\begin{table}[H]
\centering
\resizebox{\textwidth}{!}{%
\begin{tabular}{l|ccc|ccc||ccc|ccc}
  & \multicolumn{6}{c||}{\bf Rising Action} & \multicolumn{6}{c}{\bf Falling Action}\\
  & \multicolumn{3}{c|}{\bf Delatorre} & \multicolumn{3}{c||}{\bf Lehne} & \multicolumn{3}{c|}{\bf Delatorre} & \multicolumn{3}{c}{\bf Lehne} \\
\hline
\textbf{Model} &
\textbf{Acc.} $\uparrow$ & \textbf{F1} $\uparrow$ & \textbf{MSE} $\downarrow$ & \textbf{Acc.} $\uparrow$ & \textbf{F1} $\uparrow$ & \textbf{MSE} $\downarrow$ & \textbf{Acc.} $\uparrow$ & \textbf{F1} $\uparrow$ & \textbf{MSE} $\downarrow$ & \textbf{Acc.} $\uparrow$ & \textbf{F1} $\uparrow$ & \textbf{MSE} $\downarrow$ \\
\hline 
Qwen2 72B 
  & 0.94 $\pm$ 0.18 & 0.96 $\pm$ 0.12 & 2.31 $\pm$ 0.94 
  & 0.62 $\pm$ 0.10 & 0.71 $\pm$ 0.09 & 2.66 $\pm$ 0.26 
  & 0.73 $\pm$ 0.00 & 0.84 $\pm$ 0.00 & \cellcolor{yellow!25}1.66 $\pm$ 0.00 
  & 0.84 $\pm$ 0.00 & 0.90 $\pm$ 0.00 & 1.98 $\pm$ 0.00 
  \\
DeepSeek V3 
  & 0.94 $\pm$ 0.18 & 0.96 $\pm$ 0.12 & 1.83 $\pm$ 0.81 
  & 0.63 $\pm$ 0.13 & 0.74 $\pm$ 0.09 & 2.29 $\pm$ 0.18 
  & \cellcolor{yellow!25}0.86 $\pm$ 0.00 & 0.91 $\pm$ 0.00 & \cellcolor{yellow!25}1.66 $\pm$ 0.00 
  & \cellcolor{yellow!25}0.86 $\pm$ 0.00 & 0.91 $\pm$ 0.00 & 1.77 $\pm$ 0.00 
  \\
Gemma 2 27b 
  & \cellcolor{yellow!25}1.00 $\pm$ 0.00 & \cellcolor{yellow!25}1.00 $\pm$ 0.00 & \cellcolor{yellow!25}1.69 $\pm$ 0.68 
  & \cellcolor{yellow!25}0.74 $\pm$ 0.06 & 0.81 $\pm$ 0.04 & \cellcolor{yellow!25}2.02 $\pm$ 0.09  
  & \cellcolor{yellow!25}0.86 $\pm$ 0.00 & 0.92 $\pm$ 0.00 & 1.88 $\pm$ 0.00  
  & \cellcolor{yellow!25}0.86 $\pm$ 0.00 & \cellcolor{yellow!25}0.92 $\pm$ 0.00 & 2.17 $\pm$ 0.00 
  \\
Gemma 2 9b 
  & \cellcolor{yellow!25}1.00 $\pm$ 0.00 & \cellcolor{yellow!25}1.00 $\pm$ 0.00 & 1.96 $\pm$ 0.70 
  & 0.62 $\pm$ 0.08 & 0.76 $\pm$ 0.05 & 2.44 $\pm$ 0.32 
  & 0.77 $\pm$ 0.00 & 0.86 $\pm$ 0.00 & 1.83 $\pm$ 0.00 
  & \cellcolor{yellow!25}0.86 $\pm$ 0.00 & 0.91 $\pm$ 0.00 & 2.27 $\pm$ 0.00 
  \\
Llama 2 7b 
  & \cellcolor{yellow!25}1.00 $\pm$ 0.00 & \cellcolor{yellow!25}1.00 $\pm$ 0.00 & 2.77 $\pm$ 0.73 
  & 0.63 $\pm$ 0.04 &  \cellcolor{yellow!25}0.77 $\pm$ 0.03 & 3.31 $\pm$ 0.14 
  & \cellcolor{yellow!25}0.86 $\pm$ 0.00 & \cellcolor{yellow!25}0.93 $\pm$ 0.00 & 3.52 $\pm$ 0.00 
  & 0.81 $\pm$ 0.00 & 0.90 $\pm$ 0.00 & 3.20 $\pm$ 0.00 
  \\
Llama 3 70b 
  & 0.81 $\pm$ 0.37 & 0.83 $\pm$ 0.36 & 3.20 $\pm$ 0.98 
  & 0.69 $\pm$ 0.09 & 0.76 $\pm$ 0.09 & 3.13 $\pm$ 0.30 
  & 0.73 $\pm$ 0.00 & 0.81 $\pm$ 0.00 & 2.56 $\pm$ 0.00 
  & 0.79 $\pm$ 0.00 & 0.86 $\pm$ 0.00 & 2.53 $\pm$ 0.00  
  \\
Llama 3 8b 
  & 0.69 $\pm$ 0.37 & 0.75 $\pm$ 0.35 & 3.14 $\pm$ 0.40 
  & 0.64 $\pm$ 0.08 & 0.75 $\pm$ 0.07 & 2.82 $\pm$ 0.24 
  & 0.59 $\pm$ 0.00 & 0.71 $\pm$ 0.00 & 3.01 $\pm$ 0.00 
  & 0.79 $\pm$ 0.00 & 0.86 $\pm$ 0.00 & 2.48 $\pm$ 0.00 
  \\
WizardLM 2 
  & \cellcolor{yellow!25}1.00 $\pm$ 0.00 & \cellcolor{yellow!25}1.00 $\pm$ 0.00 & 2.20 $\pm$ 1.16 
  & 0.64 $\pm$ 0.10 & 0.75 $\pm$ 0.10 & 2.51 $\pm$ 0.60 
  & 0.77 $\pm$ 0.00 & 0.86 $\pm$ 0.00 & 1.90 $\pm$ 0.00  
  & \cellcolor{yellow!25}0.86 $\pm$ 0.00 & 0.91 $\pm$ 0.00 & \cellcolor{yellow!25}1.67 $\pm$ 0.00 
  \\
Mistral 7B 
  & 0.94 $\pm$ 0.18 & 0.96 $\pm$ 0.12 & 2.53 $\pm$ 0.86 
  & 0.63 $\pm$ 0.08 & 0.73 $\pm$ 0.07 & 2.60 $\pm$ 0.47 
  & \cellcolor{yellow!25}0.86 $\pm$ 0.00 & 0.92 $\pm$ 0.00 & 3.21 $\pm$ 0.00 
  & 0.84 $\pm$ 0.00 & 0.79 $\pm$ 0.00 & 2.89 $\pm$ 0.00 
  \\
Mixtral 8x7B 
  & \cellcolor{yellow!25}1.00 $\pm$ 0.00 & \cellcolor{yellow!25}1.00 $\pm$ 0.00 & 2.11 $\pm$ 0.92 
  & 0.63 $\pm$ 0.08 & 0.75 $\pm$ 0.05 & 3.01 $\pm$ 0.41 
  & 0.68 $\pm$ 0.00 & 0.77 $\pm$ 0.00 & 2.33 $\pm$ 0.00 
  & 0.72 $\pm$ 0.00 & 0.79 $\pm$ 0.00 & 1.89 $\pm$ 0.00 
  \\
\hline
Average 
  & 0.93 $\pm$ 0.13 & 0.95 $\pm$ 0.11 & 2.37 $\pm$ 0.82 
  & 0.65 $\pm$ 0.08 & 0.75 $\pm$ 0.07 & 2.68 $\pm$ 0.30  
  & 0.77 $\pm$ 0.00 & 0.85 $\pm$ 0.00 & 2.36 $\pm$ 0.00 
  & 0.82 $\pm$ 0.00 & 0.89 $\pm$ 0.00 & 2.29 $\pm$ 0.00 
  \\
\hline
\end{tabular}%
}
\caption{Brewer, and Lehne results on periods of rising or falling action when binarized the numerical scale.}
\label{tab:rising_falling_action}
\end{table}

\begin{figure*}[h]
  \centering
  \includegraphics[width=\linewidth]{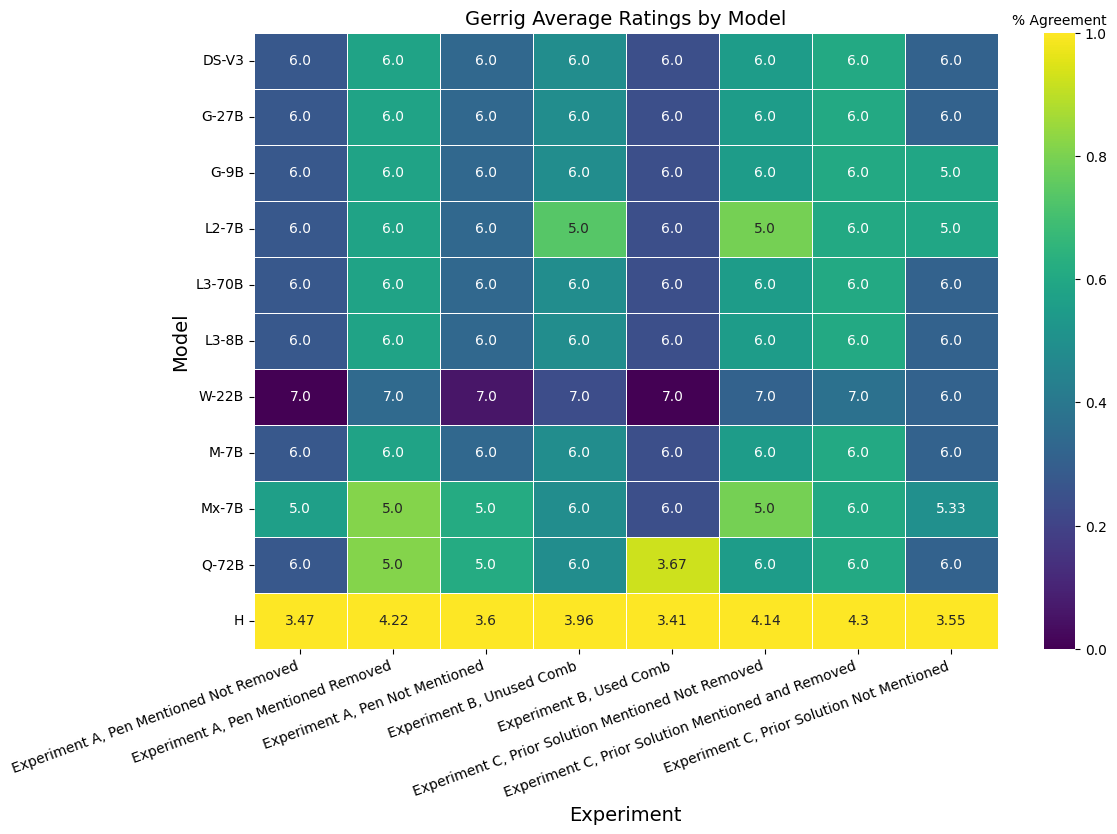}
  \captionof{figure}{Gerrig results heatmap. Rows are models. Columns are different treatments of the story.}
  \label{fig:gerrig-value}
\end{figure*}

\begin{figure}[htbp]
    \centering
    \begin{minipage}{0.5\textwidth}
        \centering
        \includegraphics[width=1\textwidth]{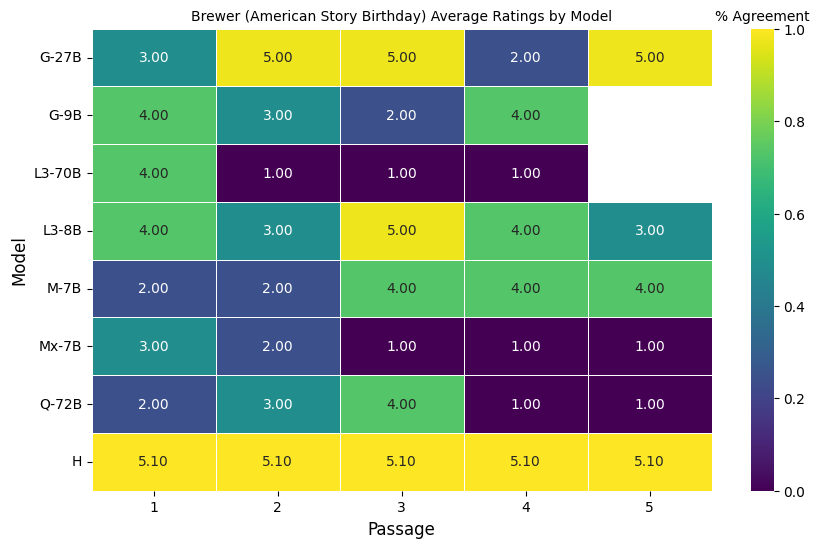}
        \subcaption[second caption.]{Brewer results (Birthday story).}\label{fig:brewer-value}
    \end{minipage}%
        \begin{minipage}{0.5\textwidth}
        \centering
        \includegraphics[width=1\textwidth]{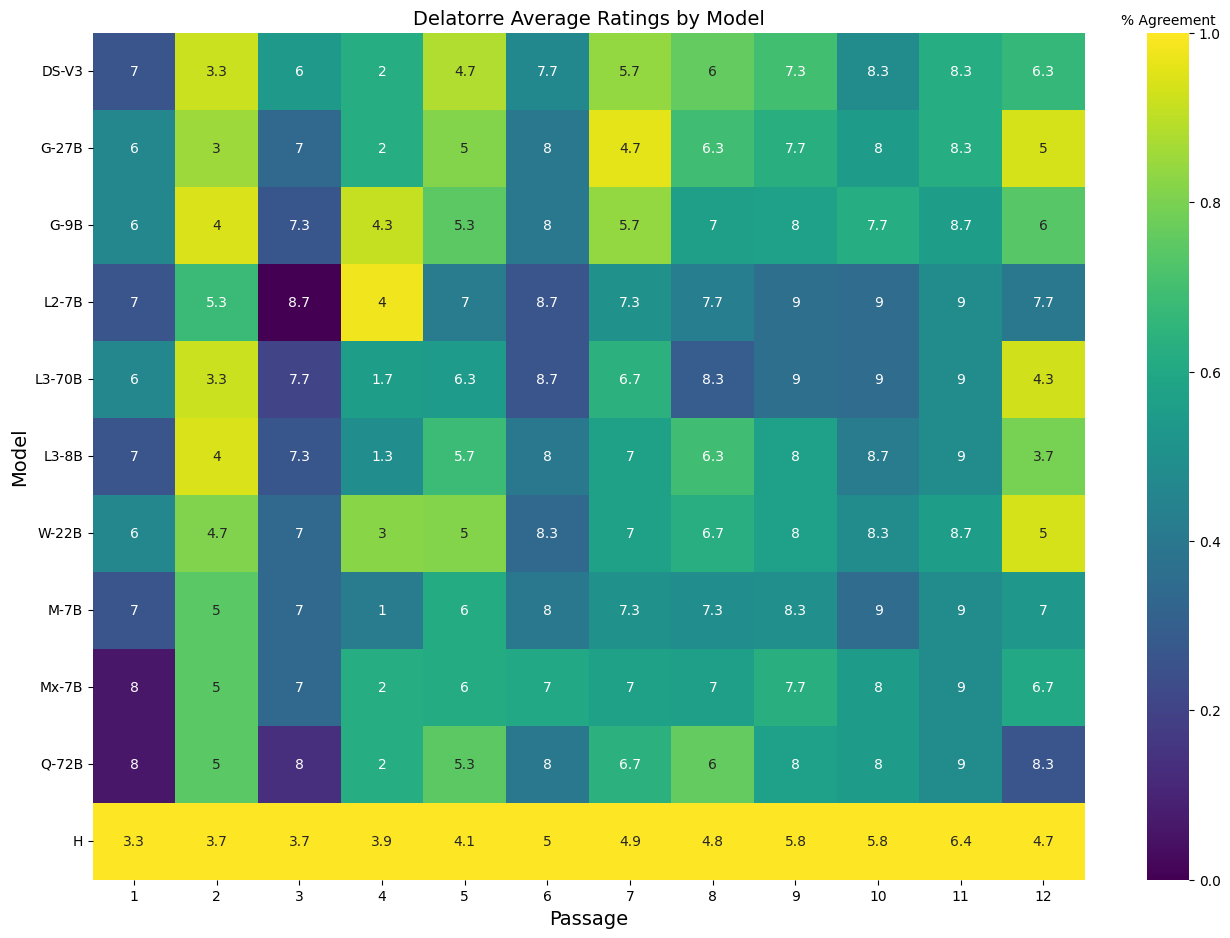}
        \subcaption[third caption.]{Delatorre results.}\label{fig:delatorre-value-appendix}
    \end{minipage}
        \begin{minipage}{1\textwidth}
        \centering
        \includegraphics[width=1\textwidth]{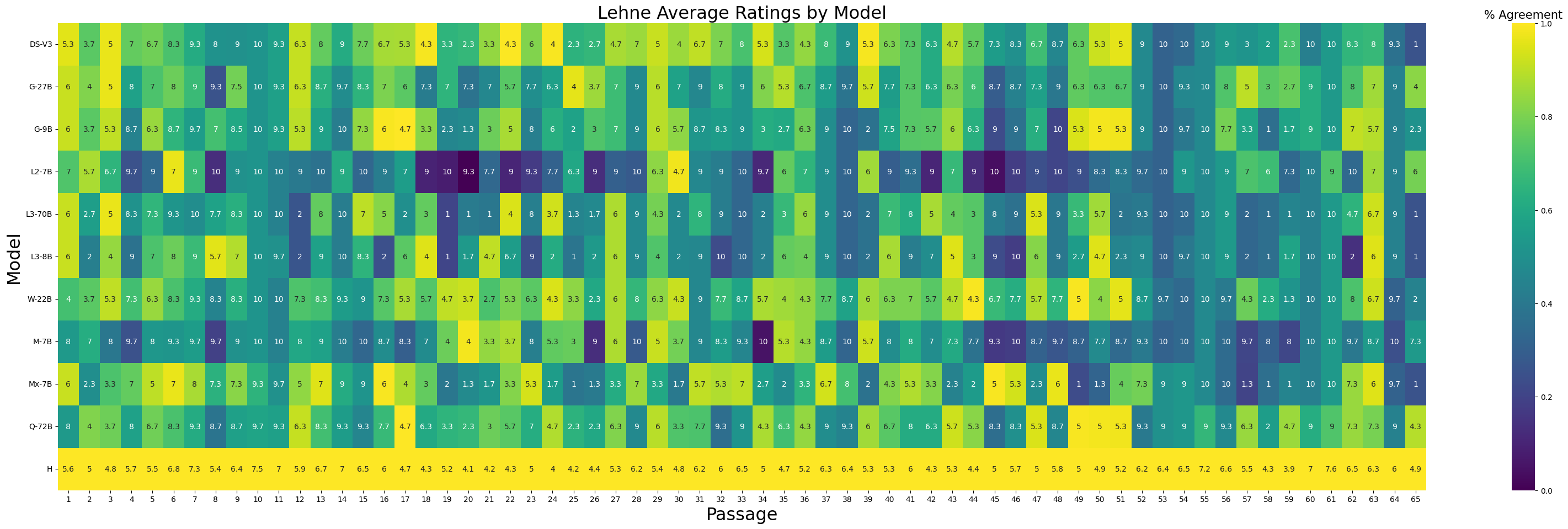}
        \subcaption[fourth caption.]{Lehne results.}\label{fig:lehne-value}
    \end{minipage}%
    \caption{Results for rating agreement. Rows are models, with bottom row the average ratings across all LMs. Columns are story segments.}
    \label{fig:enter-label}
\end{figure}

\begin{figure}[htbp]
    \centering
    \begin{minipage}{0.3\textwidth}
        \centering
        \includegraphics[width=1\textwidth]{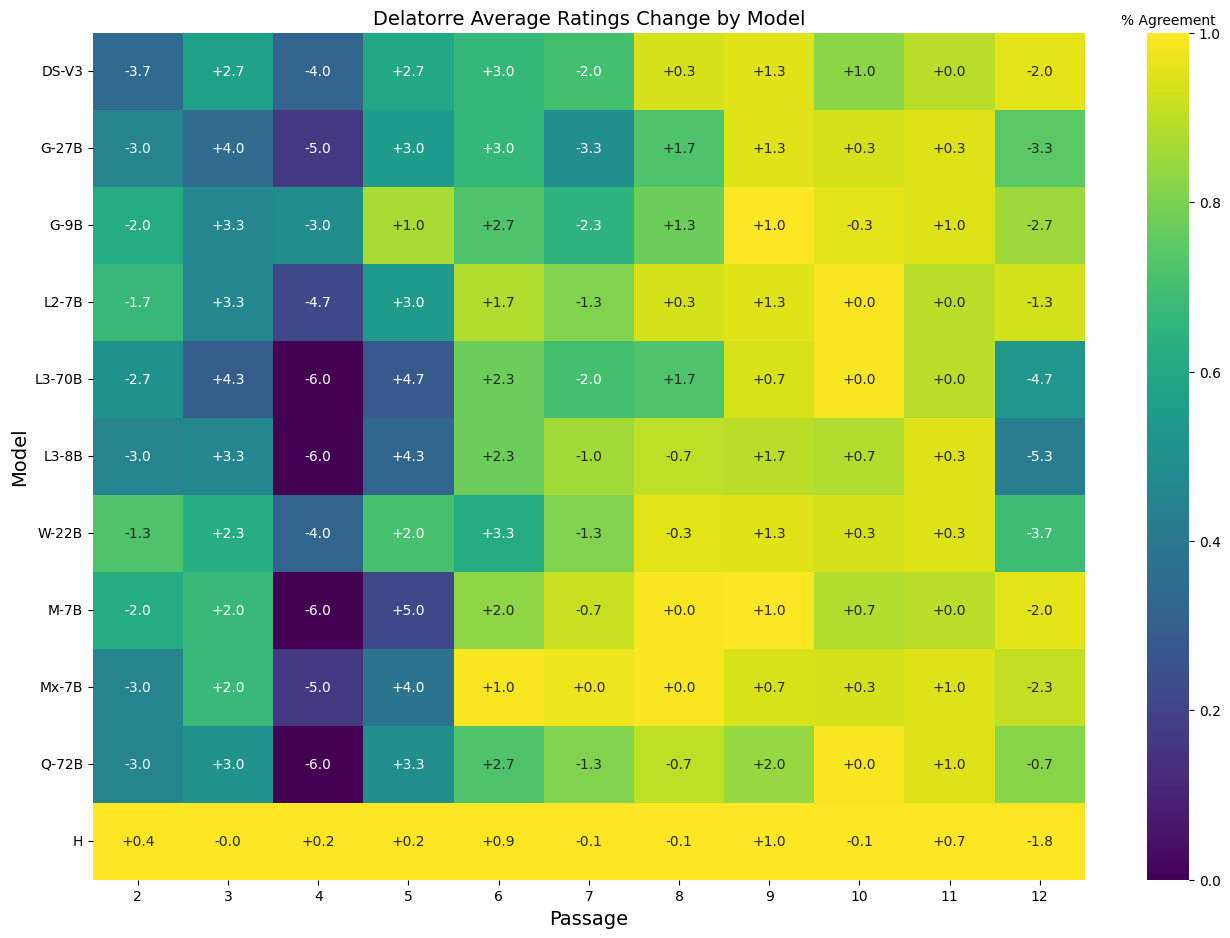}
        \subcaption[third caption.]{Delatorre  directionality.}\label{fig:delatorre-direction}
    \end{minipage}%
        \begin{minipage}{0.7\textwidth}
        \centering
        \includegraphics[width=1\textwidth]{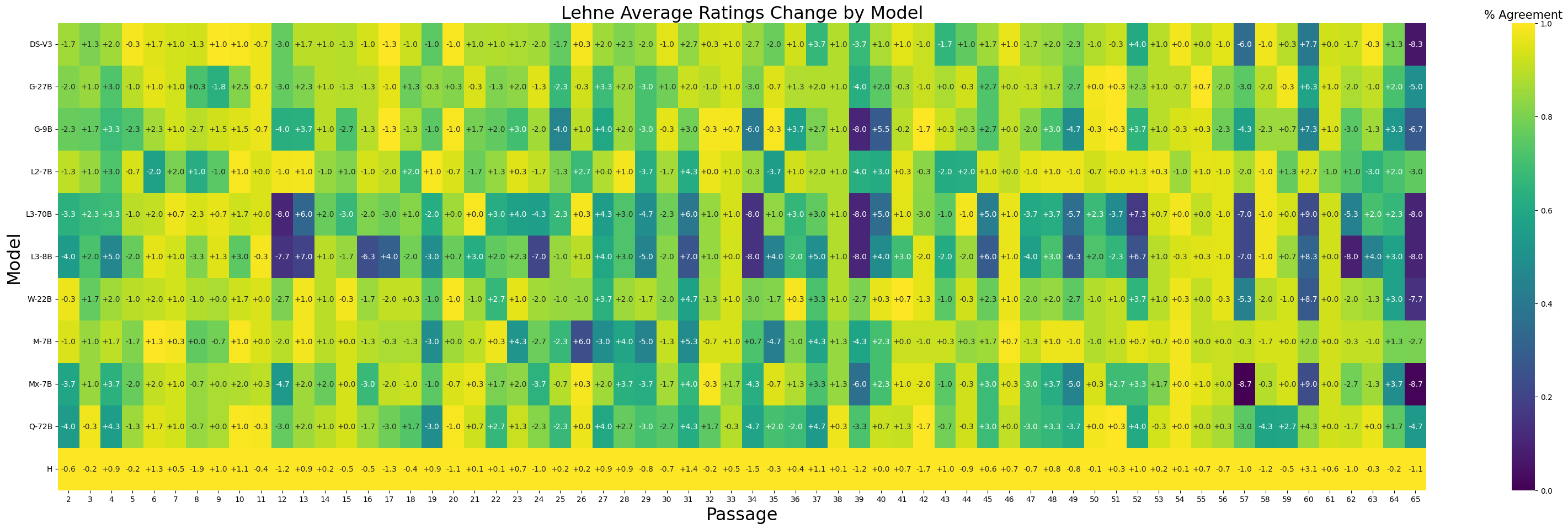}
        \subcaption[fourth caption.]{Lehne  directionality.}\label{fig:lehne-direction-appendix}
    \end{minipage}
    \begin{minipage}{0.3\textwidth}
        \centering
        \includegraphics[width=1\textwidth]{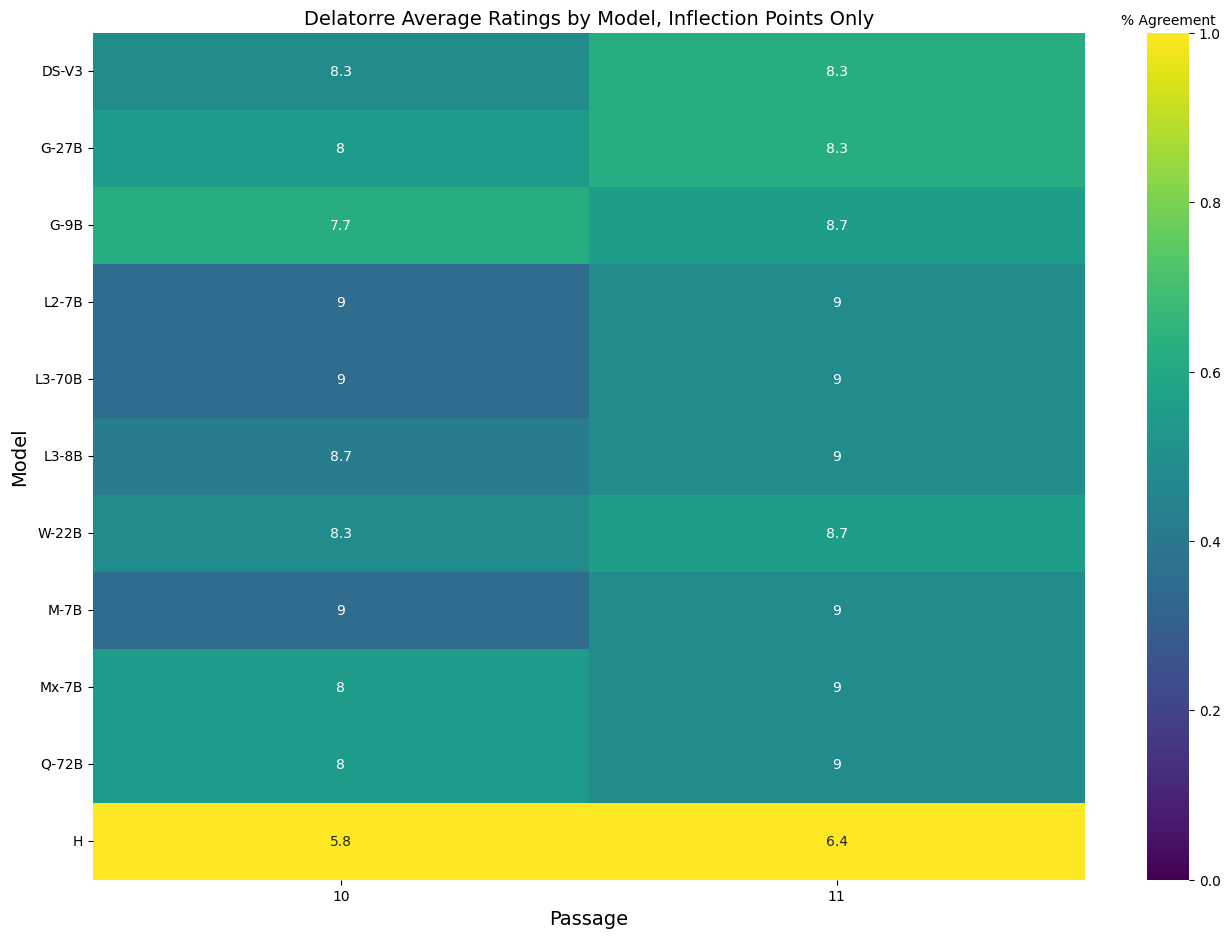}
        \subcaption[third caption.]{Delatorre inflection pts.}\label{fig:delatorre-inflection}
    \end{minipage}%
        \begin{minipage}{0.7\textwidth}
        \centering
        \includegraphics[width=1\textwidth]{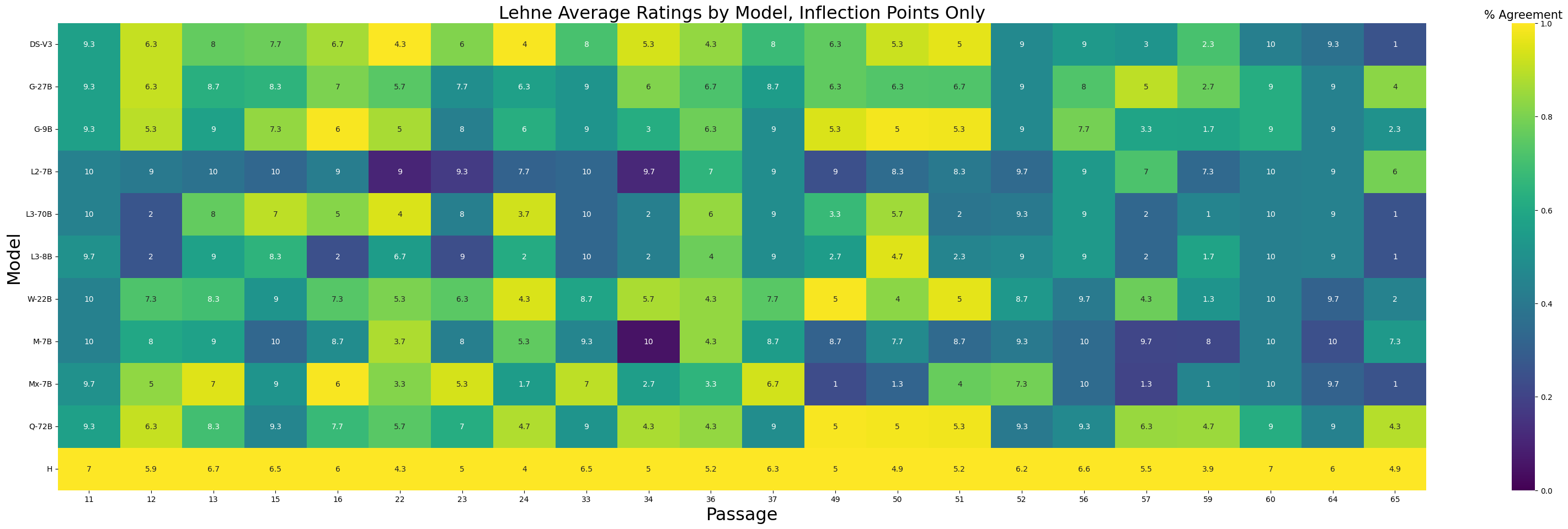}
        \subcaption[fourth caption.]{Lehne inflection points.}\label{fig:lehne-inflection-points-appendix}
    \end{minipage}
    \caption{Directionality results for Delatorre and Lehne---whether the model predicts rising or falling suspense across segments in accordance with human ratings rising or falling. Bottom graphs isolate segements where rising suspense changes to falling, and vice versa.}
    \label{fig:directionality}
\end{figure}

\begin{figure*}[h]
    \centering
    \includegraphics[width=\linewidth]{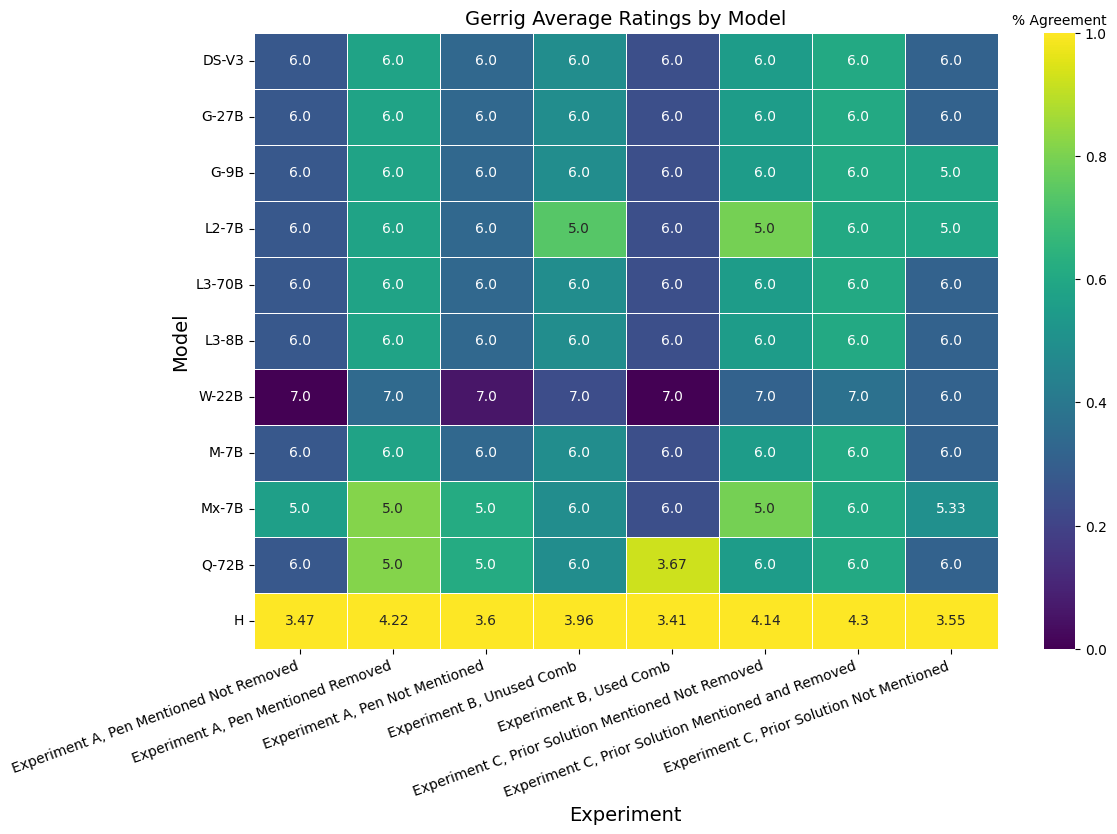}
    \caption{Gerrig average rating agreements.}
    \label{fig:gerrig-value-appendix}
\end{figure*}

\begin{figure*}[h]
    \centering
    \includegraphics[width=\linewidth]{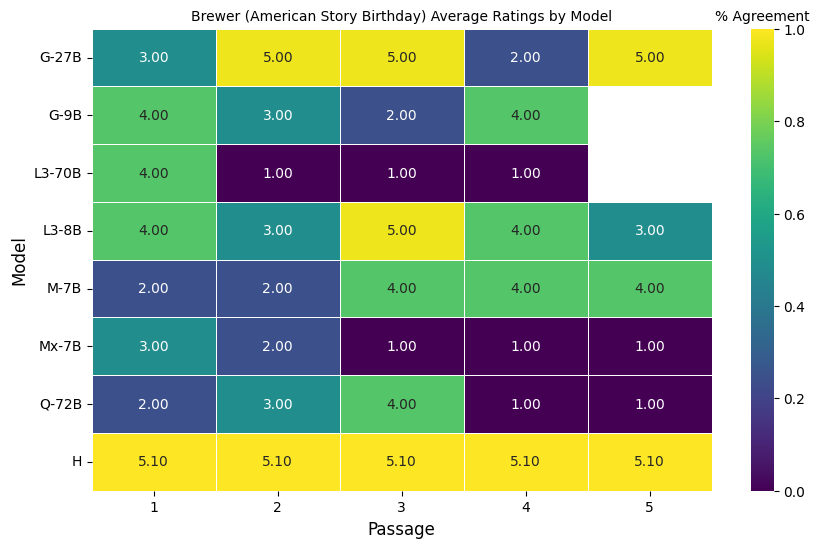}
    \caption{Brewer average rating agreements for the Birthday story.}
    \label{fig:brewer-birthday-value-appendix}
\end{figure*}

\begin{figure*}[h]
    \centering
    \includegraphics[width=\linewidth]{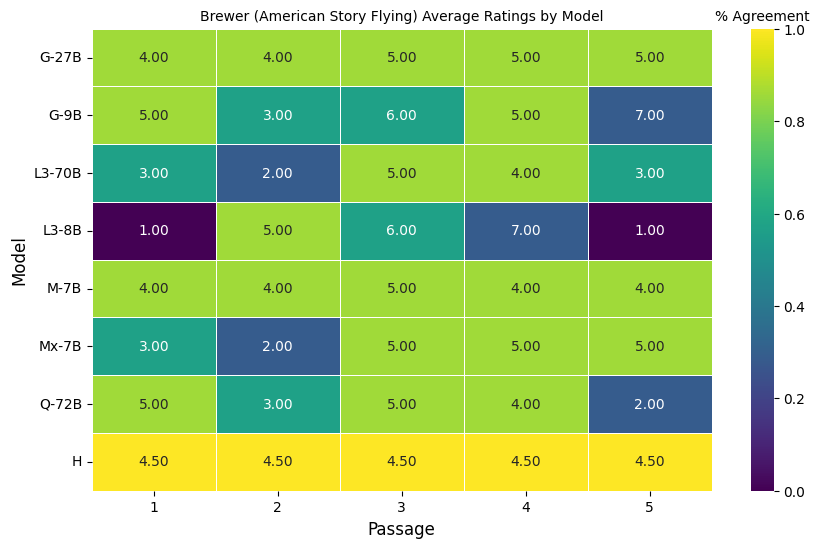}
    \caption{Brewer average rating agreements for the Flying story.}
    \label{fig:brewer-flying-value-appendix}
\end{figure*}

\begin{figure*}[h]
    \centering
    \includegraphics[width=\linewidth]{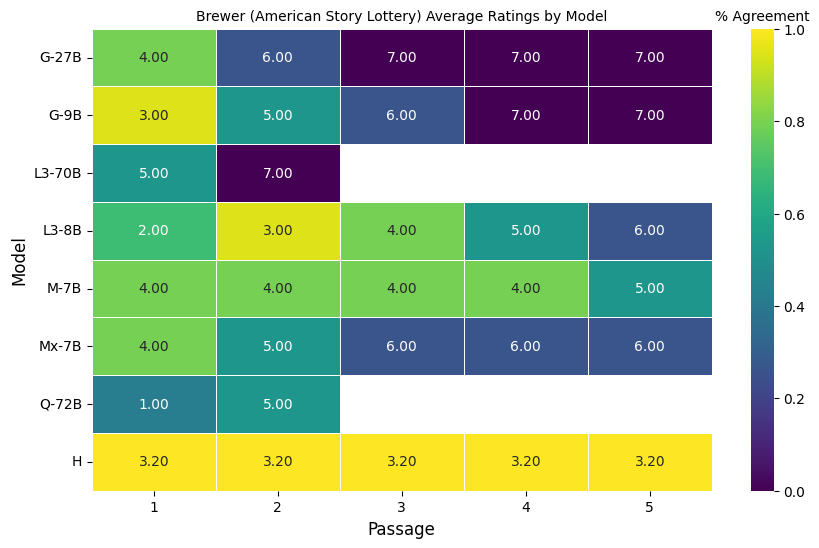}
    \caption{Brewer average rating agreements for the Lottery story.}
    \label{fig:brewer-lottery-value-appendix}
\end{figure*}

\begin{figure*}[h]
    \centering
    \includegraphics[width=\linewidth]{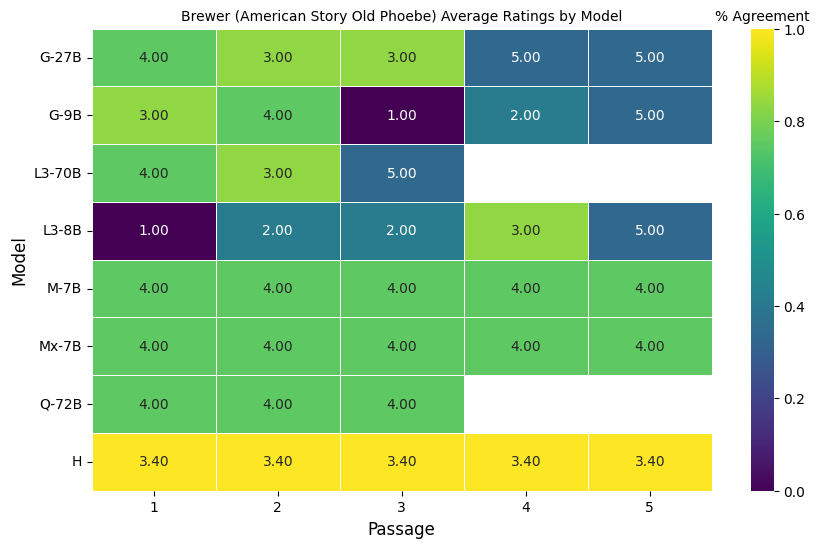}
    \caption{Brewer average rating agreements for the Old Phoebe story.}
    \label{fig:brewer-oldphoebe-value-appendix}
\end{figure*}

\begin{figure*}[h]
    \centering
    \includegraphics[width=\linewidth]{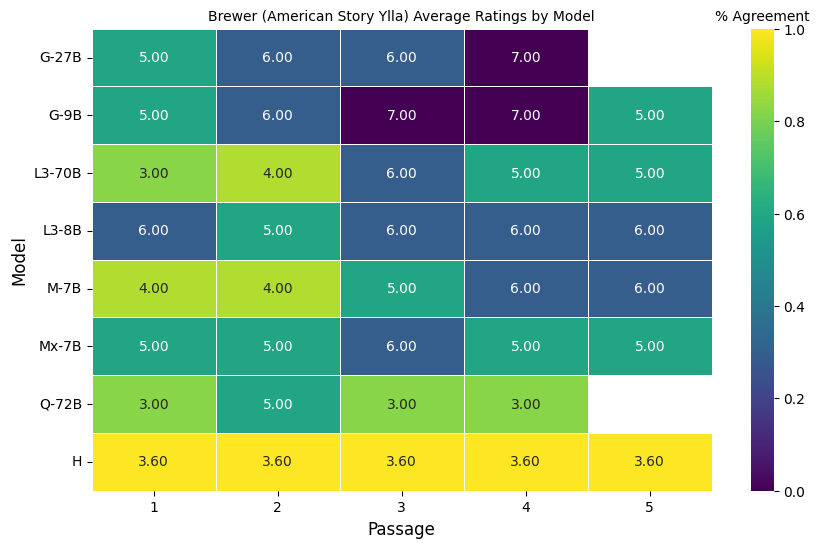}
    \caption{Brewer average rating agreements for the Ylla story.}
    \label{fig:brewer-ylla-value-appendix}
\end{figure*}

\begin{figure*}[h]
    \centering
    \includegraphics[width=\linewidth]{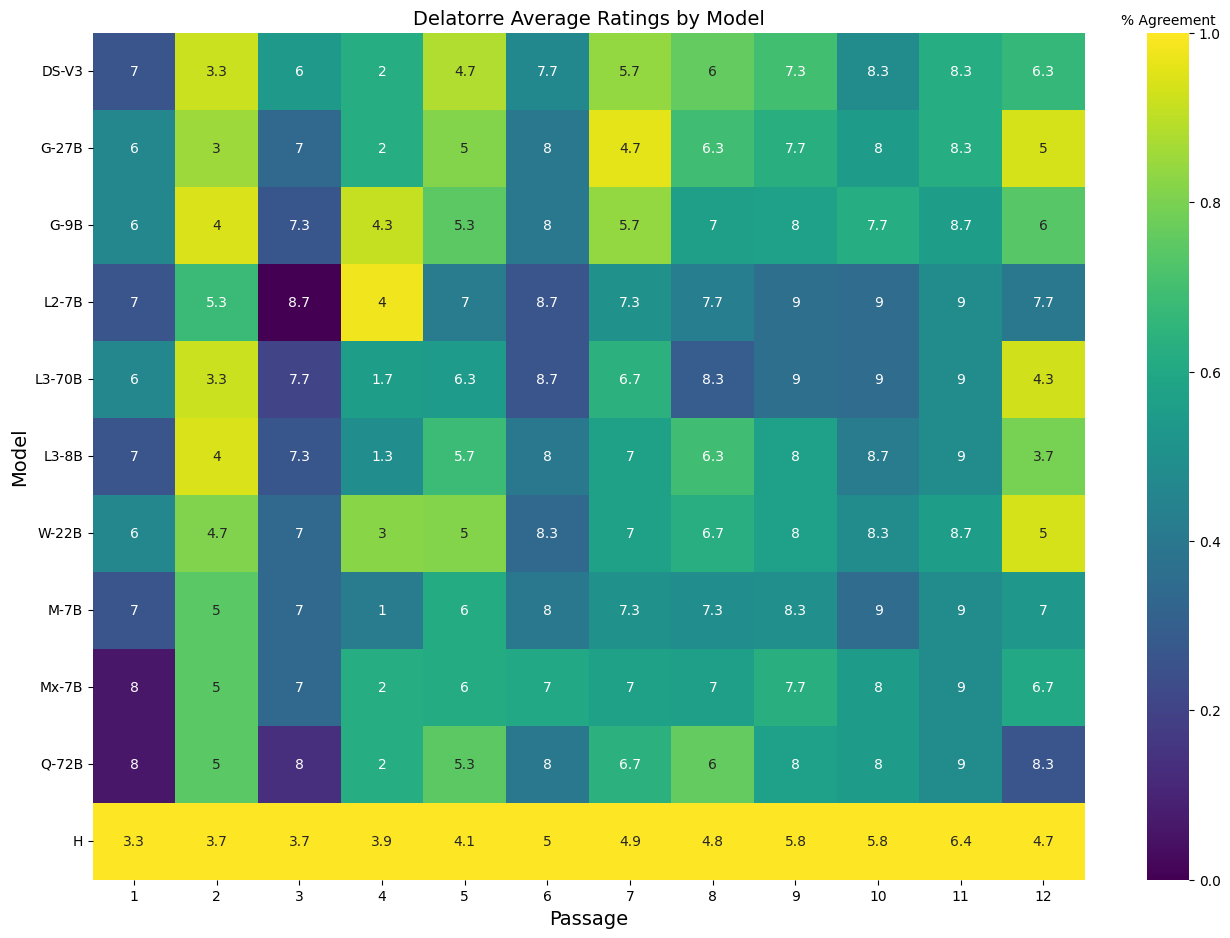}
    \caption{Delatorre average rating agreements.}
    \label{fig:delatorre-value-appendix-2}
\end{figure*}

\begin{figure*}[h]
    \centering
    \includegraphics[width=\linewidth]{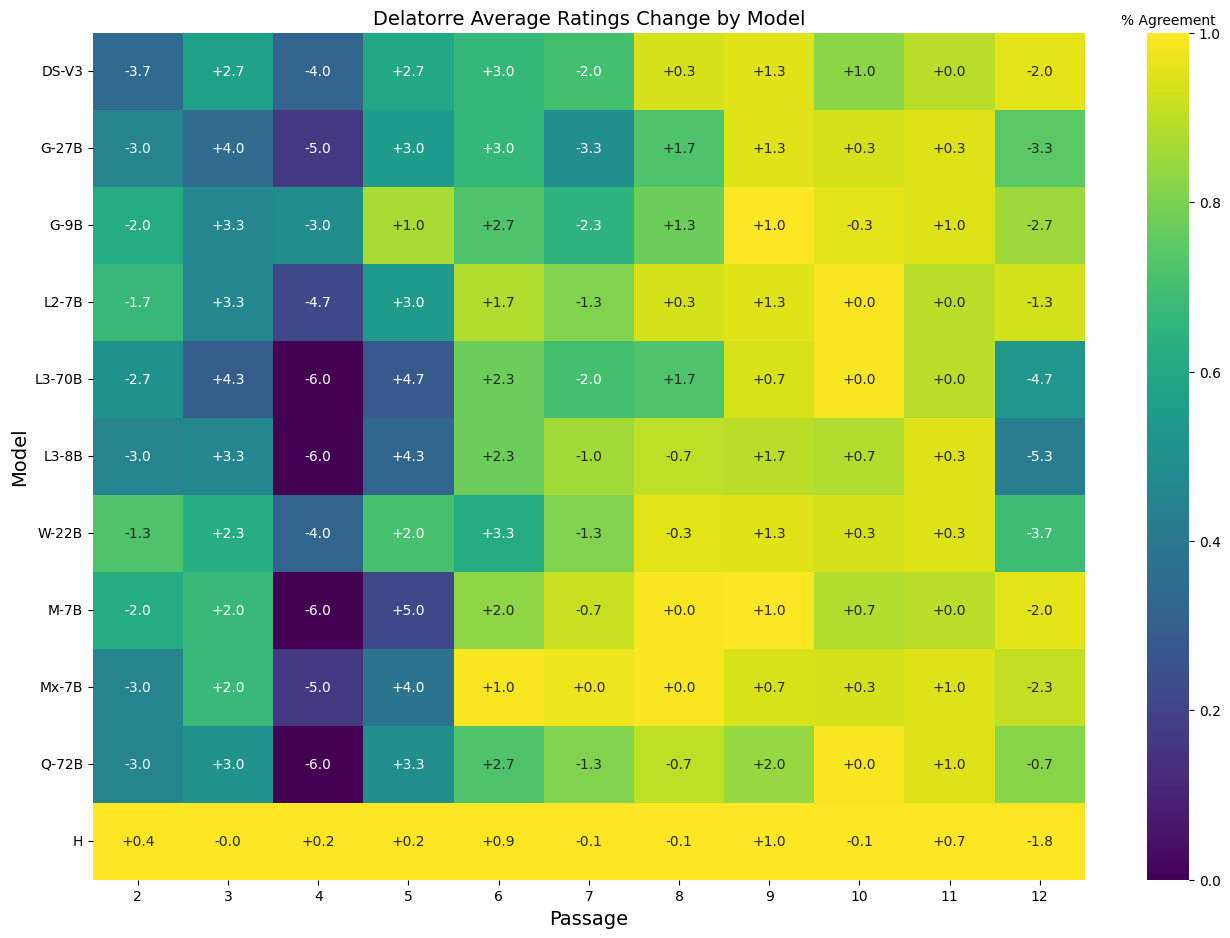}
    \caption{Delatorre average rating change agreements.}
    \label{fig:delatorre-change-appendix}
\end{figure*}

\begin{figure*}[h]
    \centering
    \includegraphics[width=\linewidth]{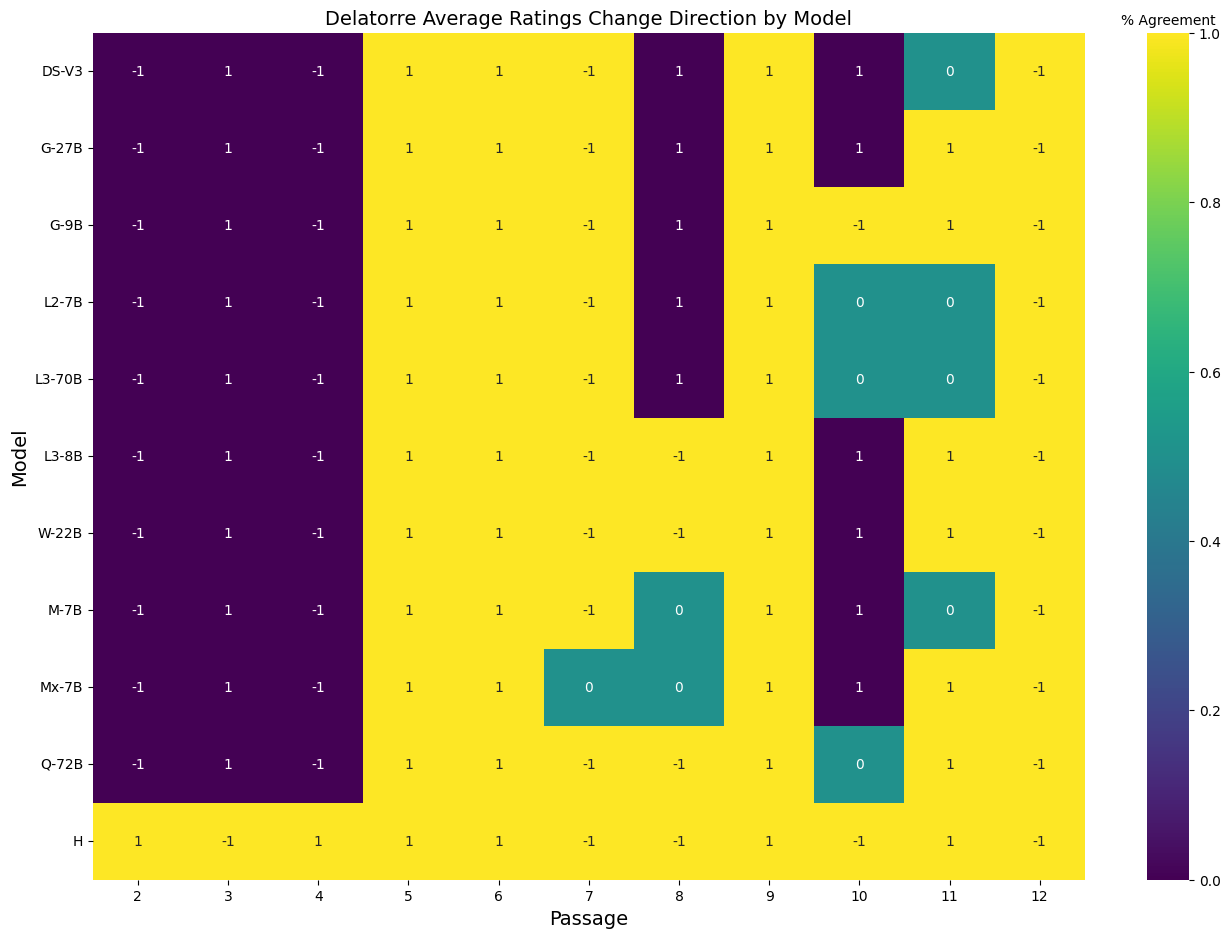}
    \caption{Delatorre average rating change direction agreements.}
    \label{fig:delatorre-change-direction-appendix}
\end{figure*}

\begin{figure*}[h]
    \centering
    \includegraphics[width=\linewidth]{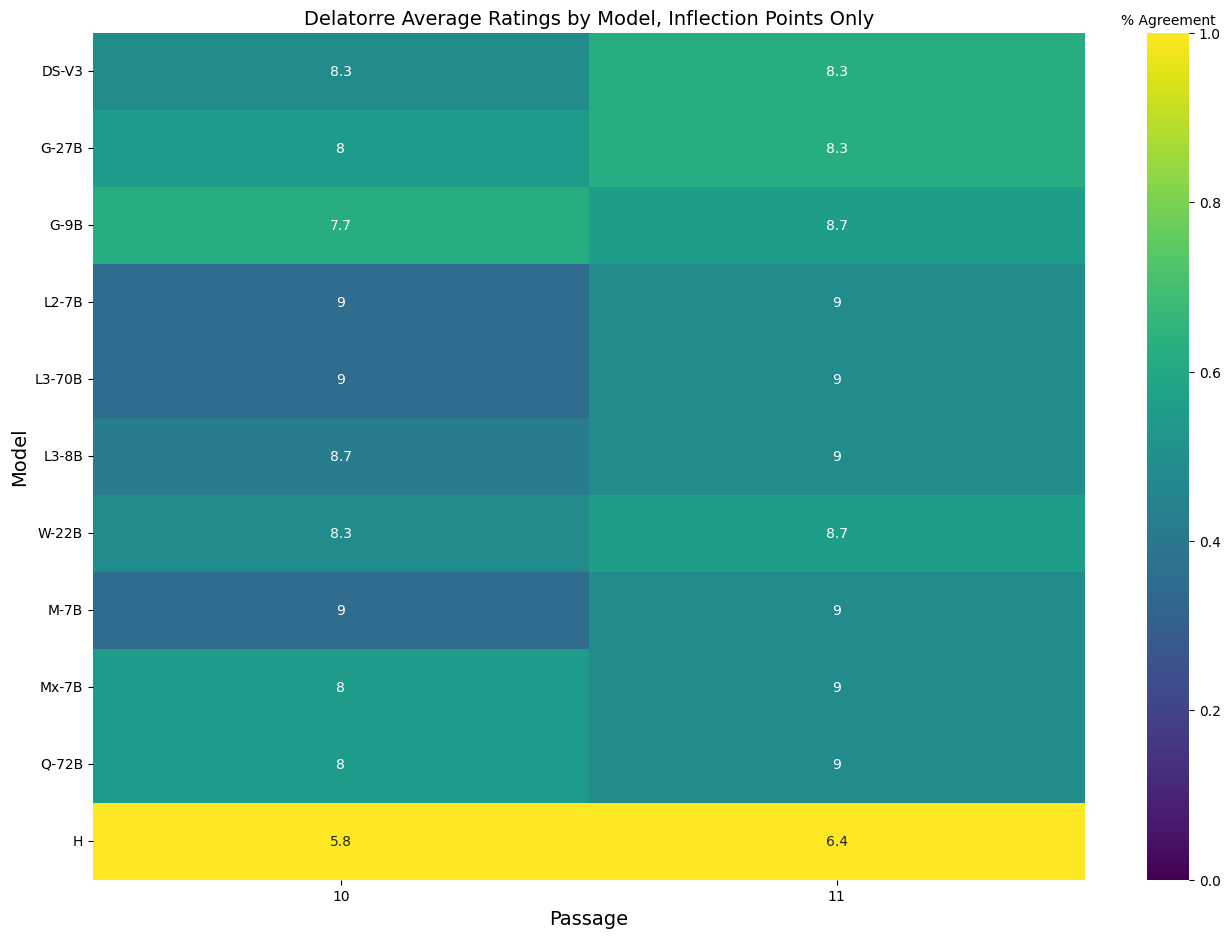}
    \caption{Delatorre average rating agreements, only passages where action changes to rising or falling.}
    \label{fig:delatorre-change-inflection-appendix}
\end{figure*}

\begin{figure*}[h]
    \centering
    \includegraphics[width=\linewidth]{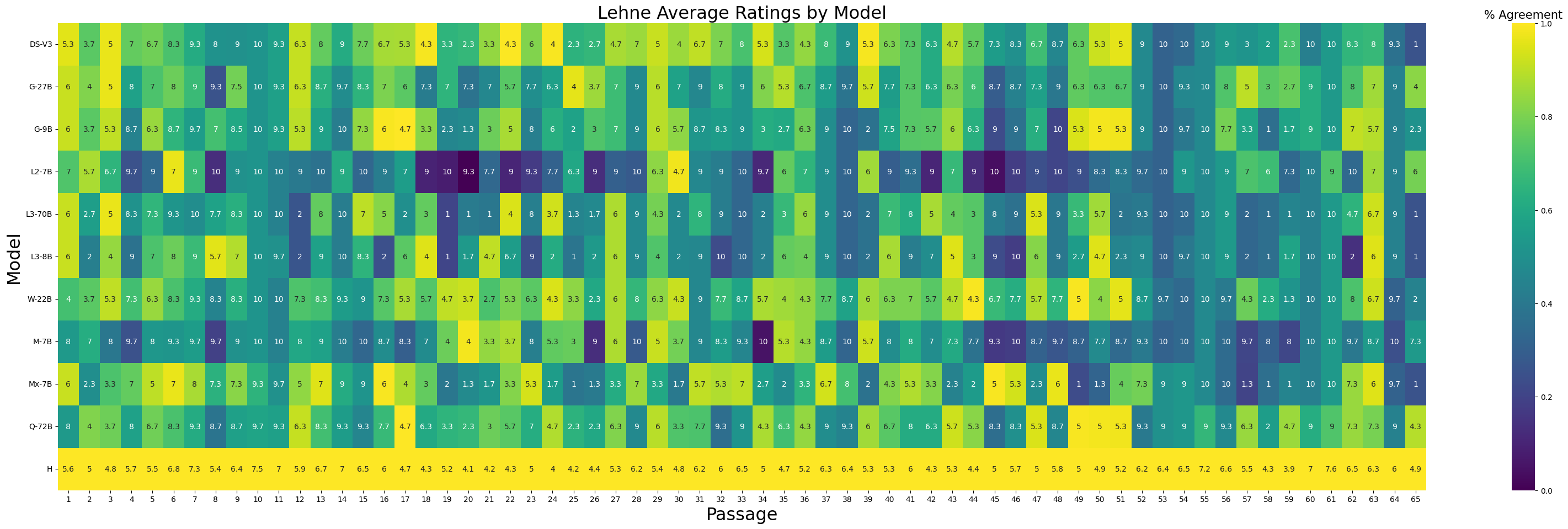}
    \caption{Lehne average rating agreements.}
    \label{fig:lehne-value-appendix}
\end{figure*}

\begin{figure*}[h]
    \centering
    \includegraphics[width=\linewidth]{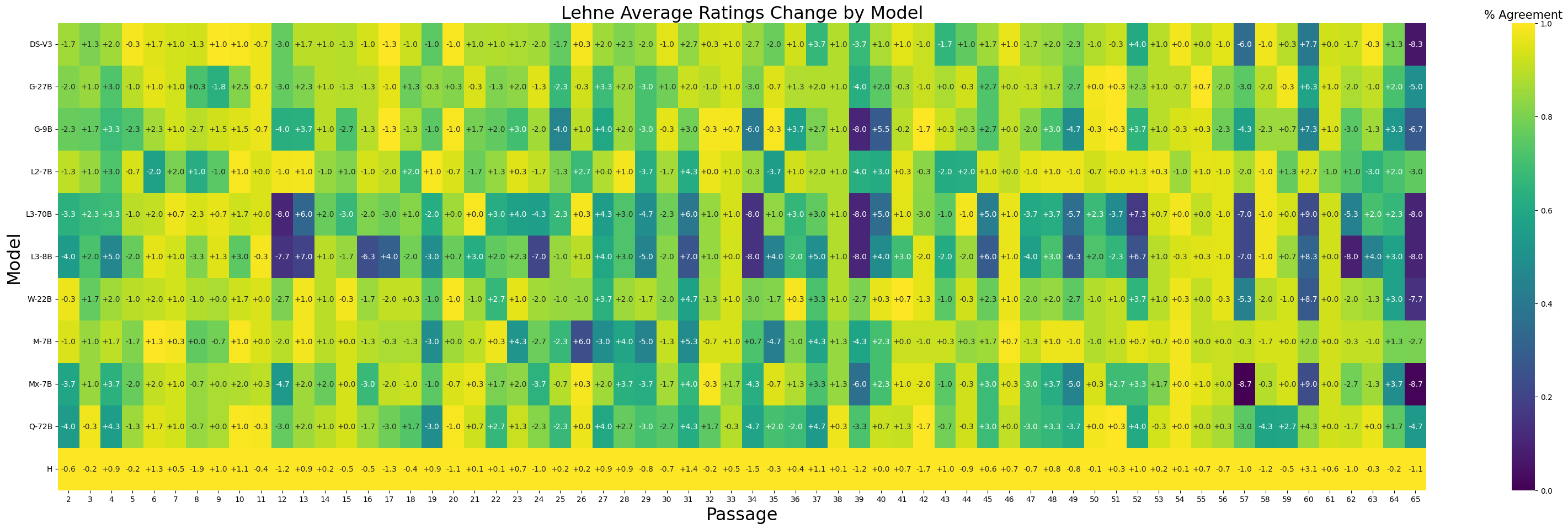}
    \caption{Lehne average rating change agreements.}
    \label{fig:lehne-change-appendix}
\end{figure*}

\begin{figure*}[h]
    \centering
    \includegraphics[width=\linewidth]{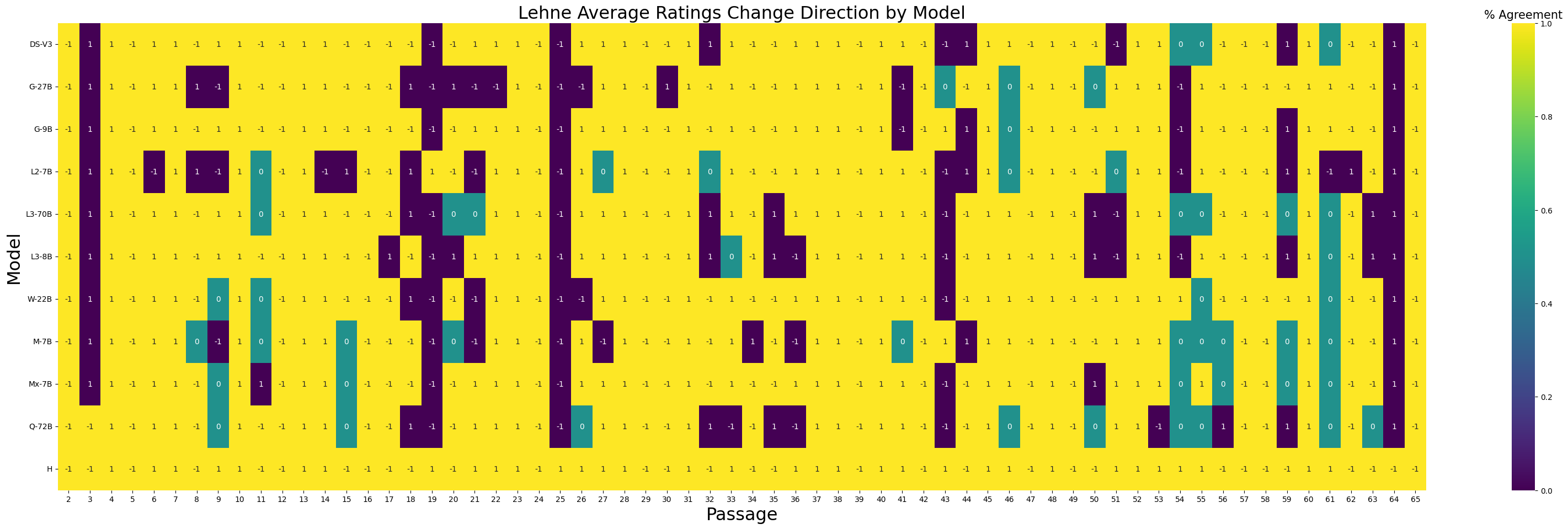}
    \caption{Lehne average rating change direction agreements.}
    \label{fig:lehne-change-direction-appendix}
\end{figure*}

\begin{figure*}[h]
    \centering
    \includegraphics[width=\linewidth]{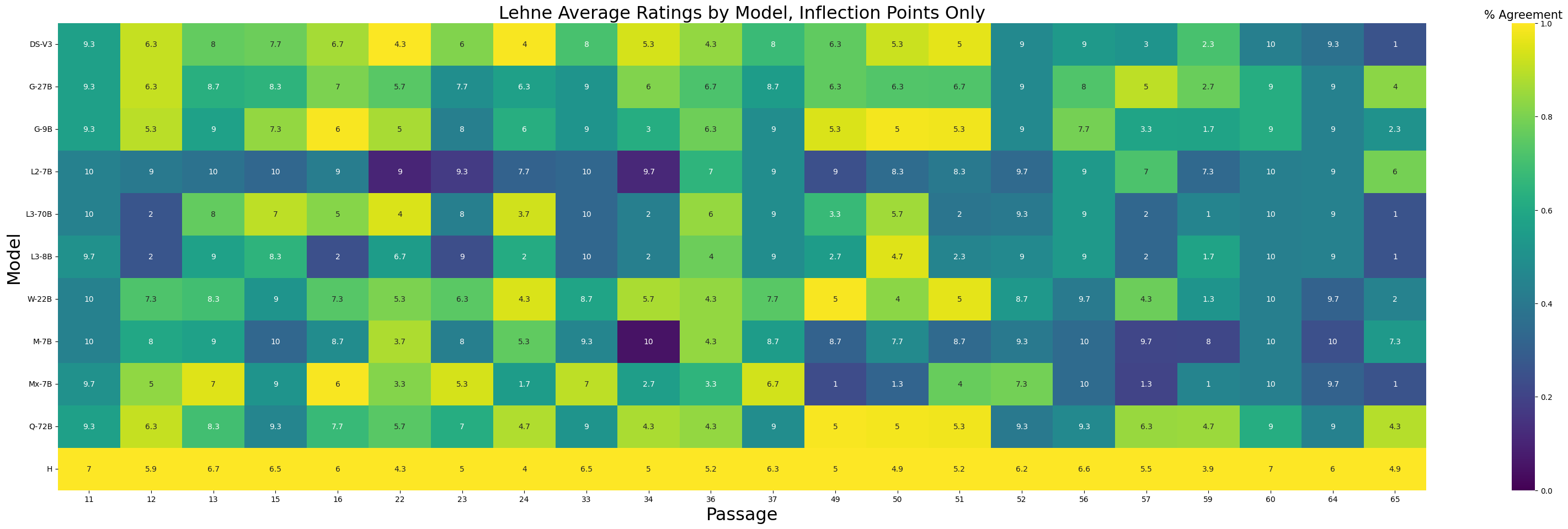}
    \caption{Lehne average rating agreements, only passages where action changes to rising or falling.}
    \label{fig:lehne-change-inflection-appendix}
\end{figure*}

\end{document}